\documentclass[default,iicol]{sn-jnl}

\usepackage{graphicx}
\usepackage{subcaption}
\usepackage{multirow}%
\usepackage{amsmath,amssymb,amsfonts}%
\usepackage{amsthm}%
\usepackage{mathrsfs}%
\usepackage[title]{appendix}%
\usepackage{xcolor}%
\usepackage{textcomp}%
\usepackage{manyfoot}%
\usepackage{booktabs}%
\usepackage{algorithm}%
\usepackage{algorithmicx}%
\usepackage{algpseudocode}%
\usepackage{array,booktabs}
\usepackage{diagbox}
\usepackage{amssymb}
\usepackage{pifont}
\usepackage{listings}%
\usepackage{lipsum}%
\usepackage{verbatim}%
\usepackage{rpm_SIunits}
\usepackage{rpm_acronyms}
\usepackage{rpm_math}
\usepackage{rpm_misc}
\usepackage{tikz} 
\usepackage{MnSymbol}
\usepackage{adjustbox}

\begin{document}

\title[Article Title]{LiDAR Odometry Survey: Recent Advancements and Remaining Challenges}


\author[1]{\fnm{Dongjae} \sur{Lee}}\email{pur22@snu.ac.kr}

\author[1]{\fnm{Minwoo} \sur{Jung}}\email{moonshot@snu.ac.kr}
\equalcont{These authors contributed equally to this work.}

\author[1]{\fnm{Wooseong} \sur{Yang}}\email{wseongy15@gmail.com}
\equalcont{These authors contributed equally to this work.}

\author*[1]{and \fnm{Ayoung} \sur{Kim}}\email{ayoungk@snu.ac.kr}

\affil[1]{\orgdiv{Department of Mechanical Engineering}, \orgname{Seoul National University}, \orgaddress{\street{1 Gwanak-ro Gwanak-gu}, \city{Seoul}, \postcode{08826}, \country{Republic of Korea}}}


\abstract{Odometry is crucial for robot navigation, particularly in situations where global positioning methods like \acf{GPS} are unavailable. The main goal of odometry is to predict the robot's motion and accurately determine its current location. Various sensors, such as wheel encoder, \acf{IMU}, camera, radar, and \acf{LiDAR}, are used for odometry in robotics. \ac{LiDAR}, in particular, has gained attention for its ability to provide rich \acf{3D} data and immunity to light variations. This survey aims to examine advancements in \ac{LiDAR} odometry thoroughly. We start by exploring \ac{LiDAR} technology and then scrutinize \ac{LiDAR} odometry works, categorizing them based on their sensor integration approaches. These approaches include methods relying solely on \ac{LiDAR}, those combining \ac{LiDAR} with \ac{IMU}, strategies involving multiple \ac{LiDAR}s, and methods fusing \ac{LiDAR} with other sensor modalities. In conclusion, we address existing challenges and outline potential future directions in \ac{LiDAR} odometry. Additionally, we analyze public datasets and evaluation methods for \ac{LiDAR} odometry. To our knowledge, this survey is the first comprehensive exploration of \ac{LiDAR} odometry.}




\maketitle

\begin{figure*}[!t]
    \centering
    \includegraphics[width=\textwidth]{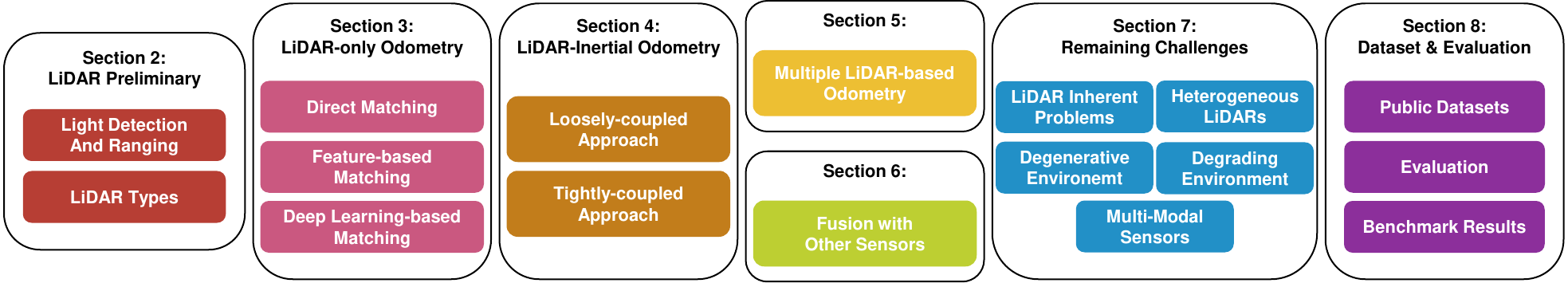}
    \vspace{-2mm}
    \caption{\textbf{Structure of \ac{LiDAR} Odometry Survey.} Section \secref{sec2} explores the intricacies of \ac{LiDAR} technology. Section \secref{sec3} -- \secref{sec6} investigate the \ac{LiDAR} odometry under different sensor modalities. Section \secref{sec7} introduces the ongoing challenges in \ac{LiDAR} odometry. Finally, Section \secref{sec8} presents the public datasets, evaluation metrics, and benchmark results.}
    \label{fig:Structure}
    \vspace{-3mm}
\end{figure*}

\section{Introduction}\label{sec1}

The history of odometry in robotics has seen a significant evolution, marked by key milestones and influential literature \cite{yousif2015overview, mohamed2019survey, jonnavithula2021lidar}. In the early stages, odometry heavily relied on wheel encoders and dead reckoning methods \cite{chong1997accurate}. However, the accuracy of wheel odometry was constrained by sensor errors stemming from wheel slippage and algorithmic inaccuracies. During this phase, researchers explored alternative approaches, shifting their focus to other sensors, such as range sensors and visual sensors. There was a concurrent surge in the field of computer vision, witnessing rapid developments in visual odometry studies \cite{scaramuzza2011visual, engel2017direct, qin2018vins}. Simultaneously, studies emerged concentrating on obtaining odometry through the use of range sensors \cite{lu1997robot, lu1997globally}, along with the advancement of scan registration algorithms such as \ac{ICP} \cite{besl1992method}. These two major research streams  --- range sensor-based odometry and visual odometry --- represent a critical juncture in the historical evolution of robotic odometry.

Further into this period, range sensors advanced, and \ac{3D} \ac{LiDAR} emerged as a transformative technology capable of measuring the surrounding space in \ac{3D}, surpassing traditional 2D measurements. Despite substantial progress, visual odometry faces limitations, particularly in low-light conditions, restricting its applicability, such as during nighttime operations. Recognizing the importance of precise location data for autonomous robots in decision-making \cite{wen2020path, sabiha2022real, jeon2022combined, huo2022dual, moon2022task, sanchez2023optimal}, researchers turned their attention to \ac{LiDAR}, which scans the surroundings in \ac{3D} while remaining unaffected by lighting conditions. This led to a rapid evolution in range sensor-based odometry using \ac{LiDAR} \cite{zhang2014loam, shan2018lego, shan2020lio}. This evolution prompts a focused review of odometry works leveraging \ac{LiDAR}.

In previous research, \citeauthor{mohamed2019survey}. \cite{mohamed2019survey} extensively reviewed approaches of odometry, placing a particular emphasis on visual-based methods. Conversely, \citeauthor{jeon2021run}. \cite{jeon2021run} presented a survey specifically tailored for unmanned aerial vehicles (UAV), focusing on the performance of visual odometry algorithms when implemented on NVIDIA Jetson platforms. Their assessment considered factors such as odometry accuracy and resource utilization (CPU and memory usage) across different Jetson boards and trajectory scenarios. \citeauthor{wang2021challenges} \cite{wang2021challenges} summarized the major applications of odometry, pointing out an expected shift towards addressing challenges in the field. Meanwhile, \citeauthor{9127855} \cite{9127855} provided a detailed review of automotive \ac{LiDAR} technologies and associated perception algorithms, exploring various components, advantages, challenges, and emerging trends in \ac{LiDAR} perception systems for autonomous vehicles. Focusing on \ac{LiDAR}-only odometry, \citeauthor{jonnavithula2021lidar}. \cite{jonnavithula2021lidar} categorized existing works into point correspondence, distribution correspondence, and network correspondence-based methodologies. They also conducted performance evaluations for \ac{LiDAR}-only odometry literature. Similarly, \citeauthor{zou2021comparative}. \cite{zou2021comparative} performed a comprehensive analysis and comparison of \ac{LiDAR} \ac{SLAM} for indoor navigation, detailing strengths and weaknesses across real-world environments. 

Notably, our review addresses a gap observed in existing surveys. While previous works have delved into specific aspects of \ac{LiDAR} odometry, none have completely covered all methodologies. Therefore, our review aims to provide a thorough examination, encompassing not only \ac{LiDAR}-only odometry but also approaches that successfully integrate other sensors for accurate \ac{LiDAR} odometry.

The structure of this survey, illustrated in Figure \ref{fig:Structure}, unfolds as follows: Section \secref{sec2} initiates an exploration of LiDAR sensors. Subsequently, we categorize \ac{LiDAR} odometry based on sensor modality and delve into each category within respective sections. Section \secref{sec3} is dedicated to methods that solely rely on \ac{LiDAR}, while Section \secref{sec4} outlines \ac{LiDAR} odometry works that integrate \ac{IMU} sensor with \ac{LiDAR}. Section \secref{sec5} provides insights into odometry employing multiple \ac{LiDAR}s. In Section \secref{sec6}, we examine the fusion of \ac{LiDAR} sensor with other sensors, such as a camera. Following this, we delve into the unresolved challenges within \ac{LiDAR} odometry. Finally, our survey concludes by discussing available public datasets and evaluation metrics, supplemented by the presentation of benchmark results. The key contributions of this paper are as follows:
\begin{itemize}
    \item Our paper offers a comprehensive review of LiDAR odometry following the progression of the technology. We categorize the review into the following sections: LiDAR preliminary, LiDAR-only odometry, LiDAR-inertial odometry, multiple LiDARs, and fusion with other sensors.
    \item Our paper explores unresolved challenges in LiDAR odometry, offering insights and directions for future research. By addressing these challenges, we aim to catalyze advancements that enhance the accuracy and robustness of LiDAR odometry.
    \item Our paper scrutinizes existing public datasets, highlighting their distinctive characteristics. Furthermore, we provide an overview of the evaluation metrics utilized in relevant studies and present benchmark results.
\end{itemize}
\section{LiDAR Preliminary}\label{sec2}

To understand the progress and challenges in \ac{LiDAR} odometry, it is essential first to grasp the basics of \ac{LiDAR} sensors. This section investigates the fundamental principles and different categories of \ac{LiDAR} sensors.

\subsection{Light Detection And Ranging}\label{subsec21}
\ac{LiDAR}, an acronym for Light Detection And Ranging, is a powerful remote sensing technology employed for measuring distances and constructing highly detailed \ac{3D} representations of objects and environments \cite{weitkamp2006lidar, khader2020introduction, survey3}. The sensing process commences with a \ac{LiDAR} system emitting laser pulses towards a designated area. When these pulses encounter obstacles, a portion of the light reflects back to the \ac{LiDAR} sensor. Measuring the time each laser pulse takes to return and leveraging the constant speed of light, \ac{LiDAR} calculates the distance to the target.

Applied systematically across large areas and synthesized into distance measurements, \ac{LiDAR} produces a point cloud --- a collection of numerous points in \ac{3D} space. These points effectively map the \ac{3D} shape and features of the area or object. In essence, \ac{LiDAR} facilitates the creation of highly detailed and accurate \ac{3D} representations of the surrounding world, proving invaluable in various fields such as geospatial mapping \cite{chase2012geospatial, elaksher2017potential}, autonomous navigation \cite{jonnavithula2021lidar, zou2021comparative}, and environmental monitoring \cite{zhao2017mobile, weibring2003versatile}.

\subsection{LiDAR Categorization}\label{subsec22}
\ac{LiDAR} can be categorized based on their distinct imaging architectures and measurement principles, as extensively discussed in previous survey \cite{survey3}. Imaging mechanisms of \ac{LiDAR} can be classified into three main categories: mechanical LiDARs, scanning solid-state LiDARs, and flash LiDARs with non-scanning architectures. Regarding measurement principles, the primary types comprise pulsed \ac{ToF}, \ac{AMCW}, and \ac{FMCW} \ac{LiDAR}s. Additionally, \ac{LiDAR}s can be further sub-classified based on attributes such as detection range, \ac{FOV}, and wavelength, as discussed in other literature \cite{survey1, survey2}. However, in this paper, we concentrate on mechanical \ac{LiDAR}s, scanning solid-state \ac{LiDAR}s, \ac{ToF} LiDARs, and \ac{FMCW} \ac{LiDAR}s, as these variants of LiDAR hold significant relevance in the context of LiDAR odometry.

\subsubsection{Imaging Mechanisms}\label{subsubsec221}
Mechanical \ac{LiDAR}s, one of the most established configurations, operate using a rotating assembly to direct a laser beam across different angles. While mechanical \ac{LiDAR} has proven reliable in measurement quality, it is subject to limitations associated with its mechanical components. These include susceptibility to degradation over time, necessitating regular maintenance to ensure optimal functionality. The inherent moving parts can also result in slower data acquisition speeds and increased vulnerability to vibrations and external shocks.

In contrast, scanning solid-state \ac{LiDAR} systems eliminate the need for mechanical rotation with diverse mechanisms. Some apply Mirror Microelectromechanical (MEMS) \cite{6714402} technology, which utilizes a stationary laser directed at the small electromechanical mirrors, adjusting the tilt angle with input voltage difference as a substitute for rotational components. Another solution is adopting an optical phased array (OPA) \cite{Heck+2017+93+107} system. OPA establishes phase modulators to modulate the wave shapes similarly to a phased array radar.

\begin{figure}[!t]
    \centering
    \subfloat[Repetitive \label{fig:repetitive}]{
		\includegraphics[width=0.41\columnwidth]{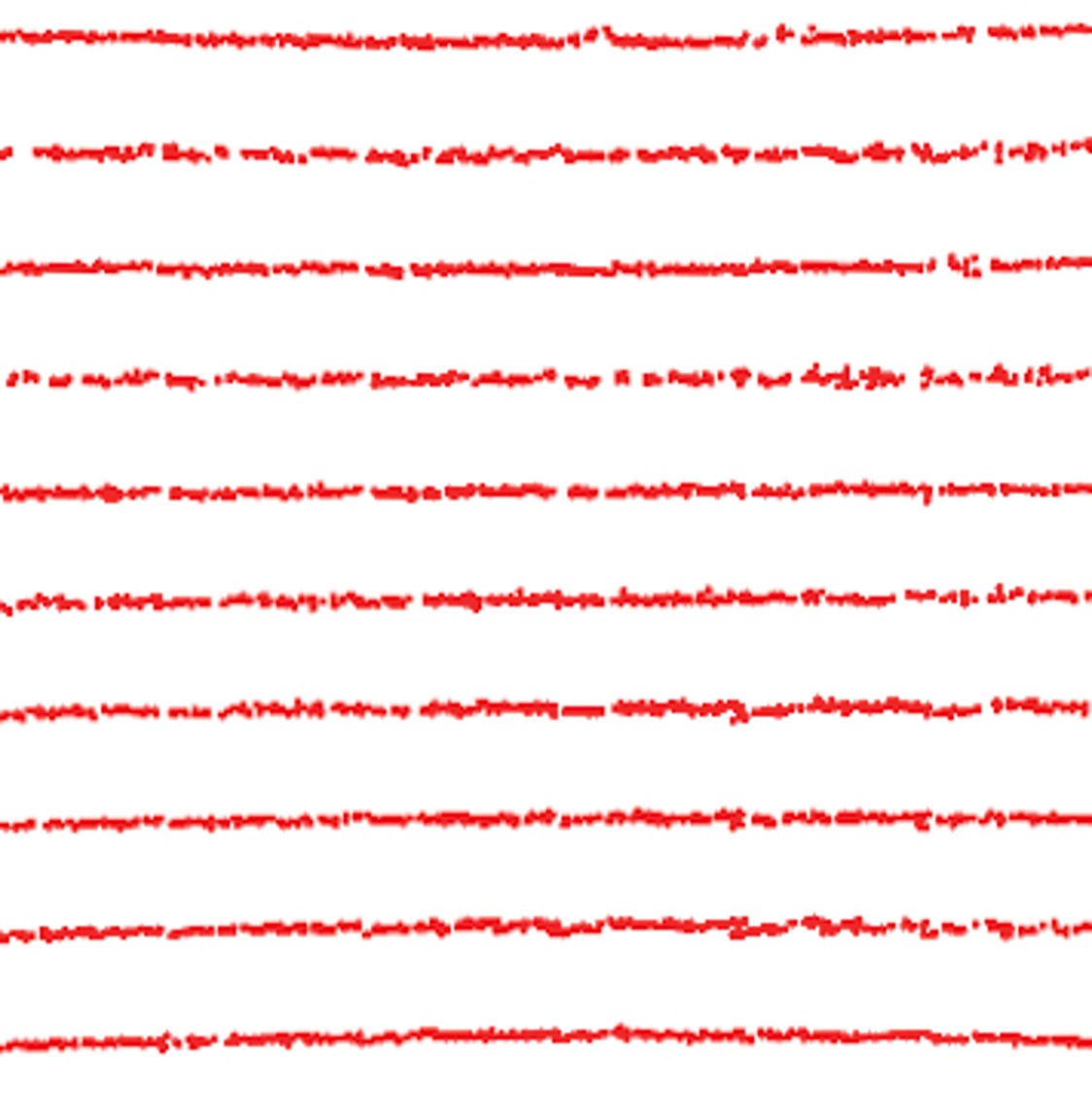}
    }
    \hspace{5mm}
    \subfloat[Non-Repetitive\label{fig:non-repetitive}]{
		\includegraphics[width=0.41\columnwidth]{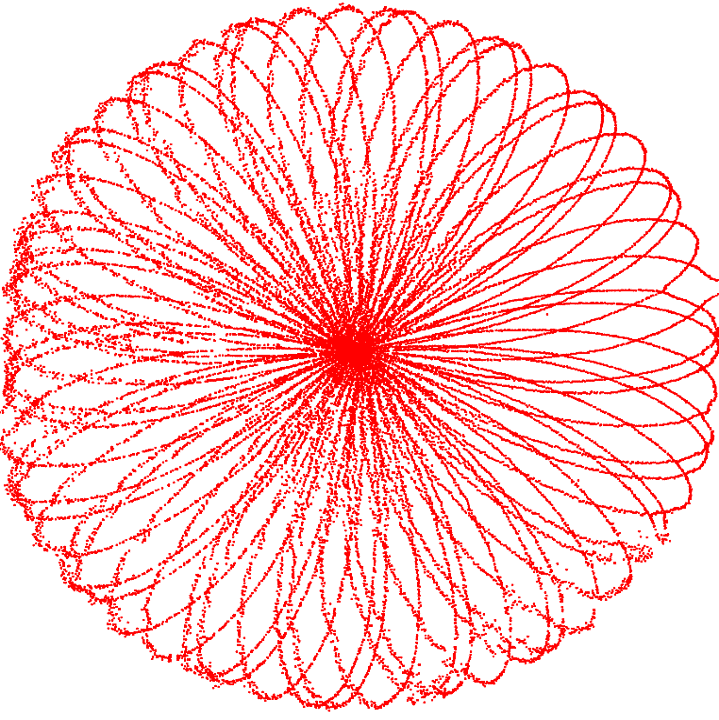}
    }
    \vspace{1mm}
    \caption{\textbf{Diverse \ac{LiDAR} Scanning Patterns.} The figure displays repetitive and non-repetitive patterns captured from real \ac{LiDAR} sensors. In (a), Velodyne VLP-16, a mechanical LiDAR, shows a vertical channel-based repetitive pattern. In (b), Livox Mid-70, a scanning solid-state LiDAR with Risely prisms, displays a unique lotus-shaped, non-repetitive pattern.}
    \label{fig:Scanning pattern}
    \vspace{-3mm}
\end{figure}

Especially, scanning solid-state \ac{LiDAR} with Risley prisms \cite{9197440} represents a notable innovation in \ac{LiDAR} community. Risley prisms allow rapid and controlled beam steering without physical movement, resulting in a more compact and robust system suitable for demanding applications. Despite the disadvantages of limited \ac{FOV}, this mitigates potential issues related to component degradation and extends the \ac{LiDAR} system's operational lifespan. Their intricate scanning patterns also ensure exhaustive environmental mapping, a critical aspect for achieving reliable \ac{LiDAR} odometry. Figure \ref{fig:Scanning pattern} visually represents distinguishing scanning patterns of \ac{LiDAR}s.

\subsubsection{Measurement Principles}\label{subsubsec222}
\ac{ToF} \ac{LiDAR} operates by emitting laser pulses and measuring the time it takes for these pulses to return after bouncing off a target. The distance to the target is calculated using the speed of light and the time the laser pulse takes. This straightforward method provides high-resolution distance measurements, making it a popular choice. However, one limitation of \ac{ToF} \ac{LiDAR} is its susceptibility to external light sources, which can reduce the \ac{SNR} \cite{koskinen1992comparison}.

On the other hand, \ac{FMCW} \ac{LiDAR} executes by continuously projecting light with a varying frequency and analyzing the frequency shift of the reflected light. This frequency shift is directly proportional to the target's distance, enabling precise distance measurements. \ac{FMCW} \ac{LiDAR} offers several notable advantages, including inherent resilience to interference due to its continuous wave signal, which helps mitigate issues caused by multi-path reflections. Moreover, \ac{FMCW} \ac{LiDAR} provides the relative velocity of the objects by analyzing the frequency shift, which proves particularly valuable in dynamic environments. However, it is important to note that \ac{FMCW} \ac{LiDAR} systems tend to be more intricate and potentially pricier compared to \ac{ToF} LiDARs.

\ac{LiDAR} technologies, each possessing unique strengths, play an integral role in \ac{LiDAR} odometry. Tailored to diverse operational needs, they can provide a range of options for capturing accurate depth data across different applications.

\section{LiDAR-Only Odometry}\label{sec3}

LiDAR-only odometry determines a robot's position by analyzing consecutive LiDAR scans. This involves the application of scan matching, a well-known technique in computer vision, pattern recognition, and robotics. LiDAR-only odometry can be classified into three types based on how scan matching is performed: (1) direct matching, (2) feature-based matching, and (3) deep learning-based matching. A summary of the LiDAR-only odometry literature is listed in \tabref{tab:LO_modified}.

\begin{table*}[!t]
\centering
\caption{The overarching summary of LiDAR-only odometry. \textbf{Direct}, \textbf{Feature}, and \textbf{Deep} represent Direct, Feature-based, and Deep Learning-based matching each.}
\label{tab:LO_modified}
\resizebox{\textwidth}{!}{%
\begin{tabular}{clcl}
\hline
\multicolumn{2}{c}{Method}                                                                                                                                           & Year & \multicolumn{1}{c}{Contributions}                                                                     \\ \hline
\multirow{12}{*}{Direct}             & ICP \cite{besl1992method}                                                                                    & 1992         & iteratively calculate closest point with point-to-point distance                                      \\
                                     & \citeauthor{chen1992object} \cite{chen1992object}                                                            & 1992         & point-to-plane ICP                                                                               \\
                                     & TrICP \cite{chetverikov2002trimmed}                                                                          & 2002         & improves ICP with trimmed squared method                                                                     \\
                                     & NDT \cite{biber2003normal}                                                                                   & 2003         & leverages normal distribution for registration                                             \\
                                     & Generalized-ICP \cite{segal2009generalized}                                                                  & 2009         & integrates point-to-point ICP and point-to-plane ICP                                                               \\
                                     & NICP \cite{serafin2015nicp}                                                                                  & 2015         & extends Generalized-ICP by incorporating surface normals                                                                                      \\
                                     & \citeauthor{hong2017probabilistic} \cite{hong2017probabilistic}                                              & 2017         & introduce probabilistic NDT representation                              \\
                                     & IMLS-SLAM \cite{deschaud2018imls}                                                                            & 2018         & IMLS representation for scan-to-map matching                                                 \\
                                     & LiTAMIN \cite{yokozuka2020litamin}, LiTAMIN2 \cite{yokozuka2021litamin2}                                     & 2021         & faster registration and modified cost function using KL divergence                                                      \\
                                     & DLO \cite{9681177}                                                                                           & 2022         & scan-to-map matching with selected keyframes using convex hull                                        \\
                                     & CT-ICP \cite{dellenbach2022ct}                                                                               & 2022        & interpolates the positions for continuous trajectory                             \\
                                     & KISS-ICP \cite{vizzo2023kiss}                                                                                & 2023         &  point-to-point ICP with adaptive thresholding                     \\ \hline
\multirow{10}{*}{Feature}            & LOAM \cite{zhang2014loam, zhang2017low}                                                                      & 2014         & extract edge and planar feature points for registration                                                                               \\
                                     & LeGO-LOAM \cite{shan2018lego}                                                                                & 2018         & leverages ground segmentation within LOAM framework                                                                            \\
                                     & SuMa \cite{behley2018efficient}                                                                              & 2018         & utilizes surface normals from surfel-based map                                                        \\
                                     & SuMa++ \cite{chen2019suma++}                                                                                 & 2019         & performs semantic ICP with semantic labels from RangeNet++ \cite{milioto2019rangenet++} \\
                                     & F-LOAM \cite{wang2021f}                                    & 2021         & emphasizes horizontal features to minimize false detections                                         \\
                                     & \citeauthor{zhou2021lidar}. \cite{zhou2021lidar}, $\pi$-LSAM \cite{9561933}                                    & 2021         & introduces plane adjustment in indoor situations                                         \\
                                     & MULLS \cite{pan2021mulls}                                                                                    & 2021         & scan-to-map multi-metric linear least square ICP                   \\
                                     & NDT-LOAM \cite{chen2021ndt}                                                                                  & 2021         & combines weighted NDT and feature-based pose refinement                                    \\
                                     & E-LOAM \cite{guo2022loam}                                                                                    & 2022         & performs D2D-NDT with geometric and intensity features                                                      \\
                                     & R-LOAM \cite{oelsch2021r}, RO-LOAM \cite{oelsch2022ro}                                                       & 2022         & extracts 3D triangular mesh features from reference object                                                      \\
                                     & \citeauthor{wang2022high}. \cite{wang2022high}                                                                & 2022         & coarse-to-fine odometry with NDT and PLICP                                                    \\ 
                                    & VoxelMap \cite{yuan2022efficient}                                                                & 2022         & leverages probabilistic plane representation and adaptive voxel construction                                                     \\
                                     
                                     \hline
\multirow{3}{*}{Deep}                & LO-Net \cite{li2019net}                                                                                      & 2019         & scan-to-scan LiDAR odometry network                                                 \\
                                     & LodoNet \cite{zheng2020lodonet}                                                                              & 2020         & select MKPs for odometry estimation  \\
                                     & \citeauthor{cho2020unsupervised}. \cite{cho2020unsupervised}                                                  & 2020         & unsupervised learning with VertexNet and PoseNet \\
\hline
\end{tabular}
}
\end{table*}

\subsection{Direct Matching}\label{subsec31}
The direct matching method directly calculates the transformation between two consecutive LiDAR scans, representing the most straightforward approach in LiDAR-only odometry. The \ac{ICP} algorithm \cite{besl1992method} is a commonly used technique for estimating this transformation iteratively by minimizing an error metric, typically the sum of squared distances between the matched point pairs. Robot odometry is derived by calculating the transformation between each pair of consecutive scans using the \ac{ICP} algorithm. However, the \ac{ICP} algorithm has drawbacks, including susceptibility to local minima, which necessitates a reliable initial guess. The algorithm is also sensitive to noise, such as dynamic objects. Additionally, its iterative nature can result in computational expense, sometimes causing prohibitively slow computation speed. Consequently, substantial efforts have been dedicated to enhancing the performance of the ICP algorithm for improved odometry. 

TrimmedICP(TrICP) \cite{chetverikov2002trimmed} enhances the conventional ICP algorithm by employing the least trimmed squares method instead of the standard least squares method. This modification improves computation speed and robustness by minimizing the sum of squared residuals for a subset of points with the smallest squared residuals. Point-to-plane ICP, introduced by \citeauthor{chen1992object} \cite{chen1992object}, refines the performance of the traditional point-to-point ICP by incorporating information about prevalent planes in real-world situations. Generalized-ICP \cite{segal2009generalized} integrates point-to-point ICP and point-to-plane ICP within a probabilistic framework, leveraging the covariance of points during the minimization step. This approach maintains the speed and simplicity of the standard ICP while demonstrating superior robustness against noise and outliers. NICP \cite{serafin2015nicp} extends Generalized-ICP by evaluating distances in 6D space, including 3D point coordinates and corresponding surface normals in the measurement vector. LiTAMIN \cite{yokozuka2020litamin} and LiTAMIN2 \cite{yokozuka2021litamin2} support faster registration through point reduction and modify the cost function of traditional ICP for robust registration.

Paired with the ICP algorithm, the Normal Distribution Transform (NDT) \cite{biber2003normal} algorithm provides an alternative that eliminates the challenging task of establishing point correspondences. The NDT algorithm aligns two point clouds by creating a normal distribution associated with the point cloud. It determines a transformation that aligns the point clouds based on the likelihood within the spatial probability function. \citeauthor{hong2017probabilistic} \cite{hong2017probabilistic} enhance the conventional NDT algorithm by introducing a probabilistic NDT representation. They assign probabilities to point samples, addressing the degeneration effect by incorporating computed covariance. Their study demonstrates that probabilistic NDT outperforms traditional NDT in odometry estimation.

Despite advancements in scan-to-scan matching algorithms, their accuracy is inherently limited. Consequently, recent LiDAR odometry works predominantly estimate the robot's pose by utilizing both scan-to-scan and scan-to-map matching. IMLS-SLAM \cite{deschaud2018imls} estimates odometry through Implicit Moving Least Square (IMLS) representation-based scan-to-map matching. DLO \cite{9681177} creates a submap for scan-to-map matching by combining point clouds from a selected subset of keyframes, including those forming the convex hull. 

Conventional LiDAR odometry typically computes discrete odometry each time a new LiDAR point cloud is received. In contrast, certain methods aim to model a continuous trajectory, emulating the continuous motion of an actual robot. CT-ICP \cite{dellenbach2022ct} accomplishes this by interpolating the positions of individual points within the LiDAR scan between the starting and ending poses. Subsequently, a continuous-time odometry estimate is obtained by registering each point through scan-to-map matching.

\subsection{Feature-based Matching}\label{subsec32}
Feature-based approaches in LiDAR-only odometry extract feature points in the LiDAR point cloud and leverage them to estimate the transformation. Utilizing only feature points instead of the entire point cloud can improve computational speed and overall performance by eliminating outliers such as noise. The main challenge with feature-based methods lies in the selection of 'good' feature points that enhance point cloud registration performance. 

LOAM \cite{zhang2014loam, zhang2017low} identifies points on sharp edges and planar surface patches by assessing local surface smoothness and matching them to estimate the robot's motion. Subsequent developments within the LOAM framework aim to improve performance by refining feature point selection. LeGO-LOAM \cite{shan2018lego} utilizes point cloud segmentation to classify points as either ground points or segmented points, ensuring accurate feature extraction. It leverages planar features from ground points and edge features from segmented points to incrementally determine a 6 \ac{DOF} transformation. R-LOAM \cite{oelsch2021r} and RO-LOAM \cite{oelsch2022ro} optimize the robot's trajectory by incorporating mesh features derived from the \ac{3D} triangular mesh of a reference object with a known global coordinate location.

Plane features, prevalent in everyday environments, have garnered significant attention as they can be easily extracted from the LiDAR point cloud. SuMa \cite{behley2018efficient} employs surface normals for odometry by comparing vertex and normal maps from the current scan with those rendered from a surfel-based map. SuMa++ \cite{chen2019suma++} integrates semantic information from RangeNet++ \cite{milioto2019rangenet++} into the surfel-based map \cite{behley2018efficient} and applies Semantic ICP, adding semantic constraints to the objective function of the ICP algorithm. F-LOAM \cite{wang2021f} emphasizes extracting distinctive horizontal features from the point cloud of mechanical LiDAR, where data is sparse vertically and denser horizontally. This approach minimizes the risk of false feature detection in the horizontal plane. \citeauthor{zhou2021lidar}. \cite{zhou2021lidar} and $\pi$-LSAM \cite{9561933} jointly optimize keyframe poses and plane parameters, referred to as plane adjustment (PA), in indoor environments. MULLS \cite{pan2021mulls} extracts diverse feature points (ground, facade, pillar, beam) and employs scan-to-map multi-metric linear least square ICP (MULLS-ICP). VoxelMap \cite{yuan2022efficient} employs adaptive-size, coarse-to-fine voxel construction for robust handling of varying environmental structures and sparse, irregular LiDAR point clouds. It addresses uncertainties from both LiDAR measurement noise and pose estimation error through probabilistic plane representation.

Instead of the variants of the ICP algorithm, the NDT algorithm can be employed, even when using features. NDT-LOAM \cite{chen2021ndt} initially obtains approximate odometry using the weighted NDT (wNDT) algorithm. This initial estimate is then refined by incorporating corner and surface features. E-LOAM \cite{guo2022loam} extracts geometric and intensity features, enhances these features with local structural information, and estimates odometry with D2D-NDT matching. \citeauthor{wang2022high}. \cite{wang2022high} propose a coarse-to-fine registration metric with NDT and PLICP (point-to-line ICP) \cite{4543181}. The roughly estimated pose with NDT serves as the initial guess for PLICP, resulting in a more accurate pose estimation.

\subsection{Deep Learning-based Matching}\label{subsec33}
While direct and feature-based methods exhibit effective performance in various environments, they often encounter difficulties with correspondence matching. It is crucial to maintain feature consistency and find the relationship between each scan to address this challenge. Some researchers investigate deep learning approaches, which hold promise in effectively addressing these issues. LO-Net \cite{li2019net} introduces a scan-to-scan LiDAR odometry network that predicts normals, identifies dynamic regions, and incorporates a spatiotemporal geometrical consistency constraint for improved interactions between sequential scans. LodoNet \cite{zheng2020lodonet} utilizes a process of back-projecting matched keypoint pairs (MKPs) from LiDAR range images into a 3D point cloud. This involves employing an MKPs selection module inspired by PointNet \cite{qi2017pointnet}, which aids in identifying optimal matches for estimating rotation and translation. \citeauthor{cho2020unsupervised}. \cite{cho2020unsupervised} exploit unsupervised learning in LiDAR odometry, utilizing VertexNet to quantify point uncertainty and PoseNet to predict relative pose between frames. The network incorporates geometrical information through estimating normal vectors and uses an uncertainty-weighted ICP loss. During supervised training, they address trivial solutions via \ac{FOV} loss.

\section{LiDAR-Inertial Odometry}\label{sec4}

\begin{table*}[!t]
\centering
\caption{The overarching summary of LiDAR-inertial odometry. \textbf{L}, and \textbf{T} represent loosely-coupled and tightly-coupled each.}
\label{tab:LIO_modified}
\resizebox{\textwidth}{!}{%
\begin{tabular}{clcccl}
\hline
\multicolumn{2}{c}{Method}                            & Year & Feature & Continuous & \multicolumn{1}{c}{Contributions}                                                                                             \\ \hline
\multirow{7}{*}{\textbf{L}}   
                            & LOAM \cite{zhang2014loam, zhang2017low}                   & 2014 & Yes     & No         & utilizes IMU for initial motion estimation and undistortion                                                                                                    \\
                            & \citeauthor{tang2015lidar}. \cite{tang2015lidar}          & 2015 & Yes     & No         & combines each odometry estimation from LiDAR and IMU                                                                                                                           \\
                            & \citeauthor{zhou2017lidar}. \cite{zhou2017lidar}          & 2017 & Yes     & No         & coarse pose estimation with INS/encoder and correction through NDT                                                                                            \\                        
                             & \citeauthor{zhen2017robust}. \cite{zhen2017robust}         & 2017 & Yes     & No         & merges prior motion model from IMU with LiDAR data                                                                                                                          \\
                             & \citeauthor{hening20173d}. \cite{hening20173d}           & 2017 & Yes     & No         & utilizes adaptive EKF with INS and LiDAR                                                                                                                  \\
                              & Lego-LOAM \cite{shan2018lego}              & 2018 & Yes     & No         & utilize IMU for initial motion estimation and undistortion                                                                                                    \\
                             & \citeauthor{yang2018robust}. \cite{yang2018robust}         & 2018 & Yes     & No         & applies graph optimization with INS and LiDAR                                                                                           \\ \hline
\multirow{36}{*}{\textbf{T}} & Zebedee \cite{6220900}                & 2012 & Yes     & No         & optimizes surface correspondence error and IMU measurement deviations                                                                   \\
                             & LIPS \cite{geneva2018lips}                  & 2018 & Yes     & Yes        & utilizes IMU preintegration factor and LiDAR plane factor                                                           \\
                             & IN2LAMA \cite{le2019in2lama}               & 2019 & Yes     & No         & formulates batch optimization with LiDAR and IMU bias factor \\
                             & \citeauthor{ye2019tightly}. \cite{ye2019tightly}           & 2019 & Yes     & No         & joint-optimization of LiDAR and pre-integrated IMU with rotational constraint                                                         \\
                             & LIO-SAM \cite{shan2020lio}               & 2020 & Yes     & No         & pose graph optimization with LiDAR and IMU preintegration factor                                              \\
                             & IN2LAAMA \cite{le2020in2laama}              & 2020 & Yes     & No         & adds UPM-based LiDAR deskewing to IN2LAMA                                                                         \\
                             & LINS \cite{qin2020lins}                  & 2020 & Yes     & No         & iESKF for faster odometry estimation                                                                                            \\
                             & \citeauthor{ding2020lidar}. \cite{ding2020lidar}          & 2020 & No      & No         & utilizes Bayesian network considering high dynamic scnearios                                                     \\
                             & KFS-LIO \cite{li2021kfs}               & 2021 & Yes     & No         & graph optimization scheme with effective feature selection                                                                              \\
                             & \citeauthor{li2021towards}. \cite{li2021towards}          & 2021 & Yes     & No         & hierarchical pose optimization with solid-state LiDAR                                                                   \\
                             & CLINS \cite{lv2021clins}                 & 2021 & Yes     & Yes        & employs continuous-time framework through cubic B-spline                                                                         \\
                             & LoLa-SLAM \cite{karimi2021lola}             & 2021 & No      & No         & achieves high-frequency odometry with LiDAR scan slicing                                                                                                           \\
                             & Fast-LIO \cite{xu2021fast}              & 2021 & Yes     & No         & introduces novel Kalman gain for Kalman filter                                                                                         \\
                             & LION \cite{tagliabue2021lion}                   & 2021 & Yes     & No         & incorporates observability metric for odometry evaluation                                                                         \\
                             & LIO-Vehicle \cite{xiao2021liovehi}            & 2021 & Yes     & No         & proposes motion constraints for ground vehicles                                                              \\
                              & \citeauthor{chen2021inertial}. \cite{chen2021inertial}       & 2021 & Yes     & No         & plane-driven submap matching with learning-based loop closure                                                                  \\
                             & \citeauthor{liu2022enhanced} \cite{liu2022enhanced}        & 2022 & Yes     & No         & exploits particle swarm filter with learning-based loop closure                                                                     \\
                             & \citeauthor{liu2022light} \cite{liu2022light}           & 2022 & Yes     & No         & proposes FG-LC-Net for learning-based loop closure                                                                                      \\
                             & \citeauthor{zeng2022mid360}. \cite{zeng2022mid360}         & 2022 & Yes     & No         & extracts feature based on single line depth variation                                                                                              \\
                             & \citeauthor{koide2022globally}. \cite{koide2022globally}      & 2022 & Yes     & No         & leverages Generalized-ICP-based cost and IMU preintegration factor                                 \\
                             & PGO-LIOM \cite{shen2022pgo}              & 2022 & Yes     & Yes        & gradient-free optimization with Monte-Carlo sampling                                                                          \\
                             & Wildcat \cite{ramezani2022wildcat}               & 2022 & Yes     & Yes        & sliding window-based continuous-time odometry framework                                             \\
                             & Fast-LIO2 \cite{xu2022fast}             & 2022 & No      & No         & improves Fast-LIO with direct fashion and ikd-Tree                                                                                                   \\
                             & Faster-LIO \cite{bai2022faster}            & 2022 & No      & No         & integrates iVox with FAST-LIO2                                                                                      \\
                             & RF-LIO \cite{qian2022rf}                & 2022 & No      & No         & removes dynamic points with scan matching and IMU preintegration                                                                           \\
                             & \citeauthor{hu2022novel}. \cite{hu2022novel}            & 2022 & No      & No         & leverages segmentation-based moving object detection                                                                        \\
                             & \citeauthor{li2022intensity}. \cite{li2022intensity}        & 2022 & Yes     & No         & introduces intensity edge feature within geometric planar feature                                                                           \\
                             & Point-LIO \cite{he2023point}             & 2023 & No      & No         & employs point-wise odometry estimation framework                                                                                            \\
                             & \citeauthor{setterfield2023feature}. \cite{setterfield2023feature} & 2023 & Yes     & No         & directly includes LiDAR feature correspondence factor                                                                                           \\
                             & RI-LIO \cite{zhang2023ri}                & 2023 & No      & No         & introduces photometric residuals from reflectivity image                                                                                \\
                             & \citeauthor{shi2023invariant}. \cite{shi2023invariant}    & 2023 & No      & No         & utilizes invariant EKF with invariant observer                                                                                               \\
                             & FR-LIO \cite{zhao2023fr}                & 2023 & No      & No         & adaptively divides LiDAR scan into multiple sub-frames                                                                                            \\
                             & DLIO \cite{chen2023direct}                  & 2023 & No      & Yes        & leverages hierarchical geometrical observer for state estimation                                                      \\
                             & \citeauthor{chen2023versatile}. \cite{chen2023versatile}      & 2023 & Yes     & No         & \begin{tabular}[c]{@{}l@{}}SE2 constrained pose estimation\end{tabular}                               \\
                             & LIMOT \cite{zhu2023limot}                  & 2023 & Yes     & No         & multi-object tracking for factor graph based dynamic objects filtering                                                          \\
                             & \citeauthor{kim2023adaptive}. \cite{kim2023adaptive}        & 2023 & Yes     & No         & proposes adaptive keyframing scheme for extreme environments                                                                           \\
\hline
\end{tabular}
}
\end{table*}

LiDAR-only odometry is computationally efficient without needing additional sensors. However, it cannot fully address the challenges detailed in Section \secref{sec7}. Therefore, recent LiDAR odometry commonly integrates LiDAR with IMU. IMU provides angular velocity and linear acceleration measurements, making it suitable for estimating coarse robot motion and enhancing pose estimation accuracy when used with LiDAR. LiDAR-inertial odometry can be branched into two categories based on how LiDAR and IMU data are fused: (1) loosely-coupled and (2) tightly-coupled. 

The loosely-coupled method independently estimates the state of each sensor, combines these states with weights, and then determines the robot's state. This approach offers high flexibility, as it estimates the state of each sensor individually. It facilitates easy adaptation to changes in the sensor system without extensive modifications to the existing framework as long as a suitable odometry module is created for the new sensor modality. Furthermore, it permits assigning weights to specific sensors, ensuring robustness in case one sensor performs sub-optimally, as the odometry can still utilize data from other sensors.

On the other hand, the tightly-coupled method utilizes measurements from all sensors concurrently to estimate the robot's state. This results in potentially more accurate odometry, as it incorporates a greater number of constraints during the odometry estimation process compared to the loosely-coupled method. However, this approach comes with a higher computational load, as all observations must be processed together. Additionally, it may be more susceptible to a loss of robustness if one sensor delivers poor-quality observations. A summary of LiDAR-inertial odometry literature is provided in \tabref{tab:LIO_modified}. In the following subsections, the specifics of these approaches are introduced.

\subsection{Loosely-coupled Approaches}\label{subsec41}
From the existing LiDAR-only methods, advancements were made with the development of LOAM \cite{zhang2014loam, zhang2017low} and LeGO-LOAM \cite{shan2018lego} by incorporating IMU sensor to correct distortions in LiDAR scans and provide initial motion estimates. Building on these improvements, \citeauthor{zhou2017lidar}. \cite{zhou2017lidar} estimate the coarse pose of the robot using INS and encoder data,  refining it with LiDAR odometry via the NDT algorithm. \citeauthor{tang2015lidar}. \cite{tang2015lidar} use the Extended Kalman Filter (EKF) to fuse independent position results from LiDAR and IMU sensors. Similarly, \citeauthor{zhen2017robust}. \cite{zhen2017robust} employ the Error State Kalman Filter (ESKF), merging the prior motion model from IMU with LiDAR-derived partial posterior information for improved robustness and accuracy. Additionally, \citeauthor{hening20173d}. \cite{hening20173d} utilize an adaptive EKF in their estimations, incorporating residuals from both INS with GPS and INS with LiDAR, facilitating further result refinement. On another front, \citeauthor{yang2018robust}. \cite{yang2018robust} opt for pose graph optimization, combining INS and LiDAR scan matching-based estimates for accurate and reliable state estimation. While loosely-coupled approaches improve accuracy over LiDAR-only methods and offer modular flexibility, they do not fully harness the synergy between sensors. This has led to increased research into tightly-coupled methods, which seek to maximize sensor integration for enhanced performance.

\subsection{Tightly-coupled Approaches}\label{subsec42}

Shifting the focus to tightly-coupled methods, this approach offers a distinct perspective on sensor fusion. Contrasting with the loosely-coupled techniques, tightly-coupled methods process data from multiple sensors in a unified framework. This integrated processing exploits the interdependencies among different sensor modalities, aiming to enhance both the accuracy and robustness of the state estimation process.

This approach begins with Zebedee \cite{6220900}, a pioneering effort in 3D LiDAR-inertial odometry. Zebedee optimizes surface correspondence error and IMU measurement deviations for odometry estimation. Initially, integrating IMU measurements directly into the factor graph posed computational challenges due to the high-frequency output of 6D pose parameters. The advent of the IMU preintegration method \cite{forster2015imu} addressed this issue by condensing hundreds of IMU measurements between keyframes into a single IMU preintegration factor. This facilitates the inclusion of each sensor measurement in the factor graph, accelerating the development of graph-based LiDAR odometry methods. 

Building upon these advancements, further innovations emerged in the field. LIPS \cite{geneva2018lips} constructs a factor graph with continuous IMU preintegration factors and 3D plane factors from LiDAR measurements, solving the graph-based optimization problem to obtain robot odometry. IN2LAMA \cite{le2019in2lama} utilizes upsampled preintegrated measurements (UPMs) \cite{le20183d} from IMU for de-skewing LiDAR scans, formulating a batch on-manifold optimization with LiDAR factor, IMU bias factor, and inter-sensor time-shift factor. Its next version, IN2LAAMA \cite{le2020in2laama}, introduces the IMU preintegration factor, similar to their previous work, IN2LAMA, but stands out by using UPMs to precisely de-skew all LiDAR measurements. While this advanced de-skewing process enhances accuracy, it may impact the real-time operation. In LIO-SAM \cite{shan2020lio}, motion estimated through IMU preintegration serves a dual purpose: de-skewing LiDAR scans and introducing a factor into the factor graph. In addition, \citeauthor{ye2019tightly}. \cite{ye2019tightly} leverages LiDAR scans and preintegrated IMU measurements for joint optimization with rotational-constrained refinement.

Further advancements in tightly-coupled methods have been made, focusing on feature selection and global optimization. KFS-LIO \cite{li2021kfs} introduces a metric for selecting the most effective subset of LiDAR features, streamlining existing graph-based methods. \citeauthor{li2021towards}. \cite{li2021towards} exploit hierarchical pose graph optimization with a novel feature extraction method of scanning solid-state LiDAR, which has an irregular scanning pattern and a metric weighting function for quantifying each LiDAR feature's residual. \citeauthor{koide2022globally}. \cite{koide2022globally} leverage GPU-accelerated voxelized Generalized-ICP matching cost factor and IMU preintegration factor. They employ a keyframe-based fixed-lag smoothing technique to estimate low-drift trajectories efficiently and create a factor graph that minimizes global registration errors throughout the map. Additionally, \citeauthor{setterfield2023feature}. \cite{setterfield2023feature} directly include feature correspondences from LiDAR measurement into a factor graph.

Unlike prior discrete-time methods, CLINS \cite{lv2021clins} employs a continuous-time framework utilizing cubic B-splines, allowing trajectory estimation at any given time by optimizing control points and knots. CLINS excels in handling asynchronous data from LiDAR and IMU sensors and managing high-dynamic scenarios with small knot distances. This makes it adept at handling point clouds with potential distortions due to different acquisition times. PGO-LIOM \cite{shen2022pgo} introduces a gradient-free optimization algorithm and a fully parallel Monte-Carlo sampling approach specifically designed to address challenges posed by nonlinear and noncontinuous problems that are difficult to handle with low-power onboard computers. They also integrate acceptance-rejection sampling \cite{accept} into feature matching cost, allowing the system to account for correct and incorrect feature matching concurrently. Wildcat \cite{ramezani2022wildcat} integrates asynchronous LiDAR and IMU measurements using continuous-time trajectory representations in a sliding-window fashion. DLIO \cite{chen2023direct} leverages the hierarchical geometrical observer instead of a filter for performance-guaranteed state estimation. Also, they propose a new coarse-to-fine approach for the continuous trajectory with a constant jerk and angular acceleration model to reduce computational overhead significantly.

As graph-based approaches progress, various factors are integrated into factor graphs to improve odometry performance. However, the increasing computational demands of such methods have led to a growing interest in approaches with lighter computational loads. Consequently, several filter-based approaches, often based on the classical Kalman filter, have emerged. LINS \cite{qin2020lins} utilizes an iterated Error State Kalman filter (iESKF) for faster odometry estimation compared to graph-based approaches. Despite attempts to enhance computational efficiency, the LINS system still faces challenges with a considerable computational load and slow processing speed, particularly when calculating the Kalman gain due to the substantial number of LiDAR measurements. FAST-LIO \cite{xu2021fast} successfully addresses this issue by introducing a novel Kalman gain formula. FAST-LIO2 \cite{xu2022fast} further improves accuracy by eliminating the feature extraction process and directly registering raw LiDAR measurements to the map. They also enhance computation speed with a data structure called an ikd-Tree. Faster-LIO \cite{bai2022faster} replaces ikd-Tree with incremental voxels (iVox) for faster search. \citeauthor{shi2023invariant}. \cite{shi2023invariant} utilize the Invariant EKF to mitigate the linearization errors inherent in EKF-based odometry, which can significantly impact estimation performance. The invariant EKF \cite{barrau2015non} demonstrates enhanced convergence and consistency compared to the standard EKF, resulting in more reliable results. Additionally, they introduce two novel methodologies: Inv-LIO1 and Inv-LIO2. Inv-LIO1 initially estimates the state through scan-to-scan matching and refines it using a mapping module. In contrast, Inv-LIO2 achieves superior accuracy with increased computation time by performing map-refined odometry through scan-to-map matching and integrating global map updates.

Advancements in graph-based and filter-based approaches have substantially enhanced the reliability of LiDAR-inertial odometry in typical environments. Moreover, methods are now specifically designed to robustly estimate odometry in complex scenarios such as dynamic and degenerative environments. \citeauthor{ding2020lidar}. \cite{ding2020lidar} exploit factor graph optimization based on a Bayesian network, considering high dynamic scenarios such as urban areas. RF-LIO \cite{qian2022rf} begins with an initial pose estimation using IMU preintegration. It utilizes the error between IMU preintegration and scan matching to create a range image and eliminate dynamic points. In addition, RF-LIO employs graph optimization to enhance pose estimation further. Similar with RF-LIO, \citeauthor{hu2022novel}. \cite{hu2022novel} leverage segmentation-based moving object detection and verification into FAST-LIO2 \cite{xu2022fast} to handle inaccurate data association in dynamic environments. LIMOT \cite{zhu2023limot} estimate poses of ego vehicle and dynamic target objects with trajectory-based multi-object tracking. By separating the dynamic and static object pose factors, the entire factor graph can simultaneously filter the dynamic objects with pose estimation. \citeauthor{kim2023adaptive}. \cite{kim2023adaptive} propose an adaptive keyframe generation scheme that considers the surrounding environment, enabling higher odometry accuracy in extreme environments. 

Furthermore, a variety of constraints and metrics have been developed to refine odometry accuracy further. LION \cite{tagliabue2021lion} incorporates an observability metric to anticipate potential declines in the quality of estimated odometry. This observability score guides the system's transition to an alternative odometry algorithm facilitated by a supervisory algorithm like HeRo \cite{HeRo}. LIO-Vehicle \cite{xiao2021liovehi} takes motion constraints of the ground vehicle to handle geometrically degraded environment by extending 2-\ac{DOF} vehicle dynamics to pre-integrated factor. \citeauthor{zeng2022mid360}. \cite{zeng2022mid360} propose a feature extraction scheme based on single line depth variation and is specifically designed for the non-uniform sampling point cloud characteristics of scanning solid-state LiDAR. \citeauthor{chen2023versatile}. \cite{chen2023versatile} leverages SE(2) constrained pose estimation for ground vehicle to solve non-SE(2) vehicle motion perturbation. \citeauthor{li2022intensity}. \cite{li2022intensity} improve feature extraction by incorporating intensity edge features within geometric planar features. They also employ multi-weighting functions based on residuals and registration consistency to assess the quality of each feature during the pose optimization process. Furthermore, RI-LIO \cite{zhang2023ri} combines two residual types in its state estimation process: photometric errors from reflectivity images and point-to-plane distances from geometric points. These images are generated using the Corrected Projection by Real Angle (CPBRA) method, addressing LiDAR laser projection biases.

Another method to enhance accuracy involves high-frequency odometry, where advancements are made through the development of techniques that improve emotion estimation by segmenting LiDAR scans. LoLa-SLAM \cite{karimi2021lola} achieves low-latency localization with a high temporal update rate by slicing LiDAR scans, ensuring sufficient measurements for accurate matching. This method is crucial for high-frequency odometry as it allows for more frequent and timely updates of the vehicle's position. On the other hand, FR-LIO \cite{zhao2023fr} deals with an aggressive motion by adaptively dividing LiDAR scan into multiple sub-frames, enhancing estimation robustness. Such division is essential for maintaining accuracy in high-frequency odometry, particularly in dynamic environments.  Additionally, \citeauthor{zhao2023fr}. introduce the iterated ESKFS to mitigate potential degeneration issues caused by increased sub-frames. Point-LIO \cite{he2023point} achieves high-frequency odometry through a point-by-point framework. This approach involves processing LiDAR scans at the individual point level, a strategy that naturally eliminates motion distortion. These high-frequency methods offer a path to more responsive and accurate odometry in rapidly changing scenarios.

Similar to LiDAR-only odometry, deep learning methods play a pivotal role in enhancing odometry estimation, showcasing advancements in this domain. \citeauthor{chen2021inertial}. \cite{chen2021inertial} integrates factor graph for state estimation and plane-driven submap matching with a learning-based point cloud network for loop detection. \citeauthor{liu2022enhanced} \cite{liu2022enhanced} exploits the adaptive particle-swarm filter with an efficient resampling strategy to tackle the environment diversity integrating with lightweight learning-based loop detection. \citeauthor{liu2022light} \cite{liu2022light} propose FG-LC-Net \cite{liu2020fg} for learning-baed loop closure and data structure S-Voxel to improve the speed of the system.
\section{Multiple LiDARs}\label{sec5}

The LiDAR-inertial odometry, discussed in Section \secref{sec4}, showcases impressive accuracy. Nevertheless, limited \ac{FOV} in certain LiDAR systems poses challenges to state estimation, hindering further advancements. Additionally, interference from other sensors can obscure regions within the LiDAR's \ac{FOV}. Irregular scanning patterns, observed in some scanning solid-state LiDARs, further pose challenges in achieving precise scan registrations due to sparsity.

To tackle challenges associated with single LiDAR systems, researchers are increasingly exploring the use of multiple LiDARs in odometry. Multiple LiDARs offer broader scanning coverage, reducing interference from additional sensors. Integrating diverse scanning patterns from multiple LiDARs enhances accuracy in scan registrations, surpassing reliance on a single LiDAR with a non-repetitive scanning pattern.

Pioneering research in the domain of multiple \ac{LiDAR}s-based odometry begins with M-LOAM \cite{9444284}. Assuming the synchronization of all \ac{LiDAR}s, M-LOAM involves feature extraction from each \ac{LiDAR}, data aggregation, and estimation of the robot's state. However, synchronizing multiple \ac{LiDAR}s using PPS (Pulse Per Second) introduces complexity and necessitates additional hardware requirements. On the other hand, synchronization through PTP (Precision Time Protocol) primarily aims to unify time standards but may demand extra effort to attain synchronized data. \citeauthor{lin2020decentralized}. \cite{lin2020decentralized} employ a decentralized \ac{EKF} that concurrently runs multiple \ac{EKF} instances, one for each \ac{LiDAR}. While this method can handle asynchronous \ac{LiDAR}s, it doesn't fully leverage the combined measurements from all \ac{LiDAR}s simultaneously, which reduces the benefits of using multiple \ac{LiDAR}s.

In the case of independently utilizing measurements from each \ac{LiDAR} for state estimation, the occlusion experienced by a single LiDAR can have a cascading impact on subsequent state estimation. LOCUS \cite{9293359}, which assumes that all the \ac{LiDAR}s are synchronized, points out that significant time discrepancies can result in failures in state estimation. In their subsequent research \cite{reinke2022locus}, they address this challenge by discarding delayed scans to enhance robustness, although this approach comes at the expense of losing some information. Similarly, M-LIO \cite{das2023m} acknowledges the asynchrony among \ac{LiDAR}s through signal association. However, it lacks a method to compensate for the temporal discrepancies arising from the asynchrony.

To overcome these issues, researchers have integrated \ac{IMU} sensors for correcting temporal discrepancies in asynchronous \ac{LiDAR} measurements \cite{nguyen2021miliom, wang2022simultaneous, nguyen2023slict, jung2023asynchronous}, similar to their role in \ac{LiDAR}-inertial odometry. \citeauthor{nguyen2021miliom}. \cite{nguyen2021miliom} and \citeauthor{wang2022simultaneous}. \cite{wang2022simultaneous} employ \ac{IMU} propagation to compensate for temporal discrepancies among multiple \ac{LiDAR}s. They extract edge and planar features from each point cloud and transform these features into a common reference frame aligned with the most recent acquisition time from all \ac{LiDAR}s. While these approaches successfully estimate robot trajectories, they also introduce additional challenges. \ac{IMU} propagation, which is inherently discrete due to its frequency, requires additional linear interpolation, potentially leading to additional errors. Moreover, as time discrepancies become more pronounced, the duration required to accumulate the point clouds increases, which further intensifies the dependence on \ac{IMU} for state propagation. However, the accuracy of the \ac{IMU} propagation deteriorates over extended periods due to noise, which can adversely impact the odometry. 

In addressing the challenge of discrete \ac{IMU} propagation, MA-LIO \cite{jung2023asynchronous} adopts B-spline interpolation \cite{sommer2020efficient} as an alternative for linear interpolation, effectively compensating for temporal discrepancies. Furthermore, \citeauthor{jung2023asynchronous}. \cite{jung2023asynchronous} leverage point-wise uncertainty to assign penalties based on the acquisition time, addressing the challenge of degraded \ac{IMU} propagation accuracy. On the other hand, SLICT \cite{nguyen2023slict} interprets the point clouds of each \ac{LiDAR} as a continuous stream. Combining only the point clouds captured within a designated interval, SLICT maintains a consistent accumulation duration, even when significant time discrepancies exist.

Utilizing multiple \ac{LiDAR}s for odometry addresses the limitations associated with single \ac{LiDAR} configuration, leading to improved performance. However, challenges such as optimizing \ac{LiDAR} placements \cite{hu2022investigating}, increased computational demands, and inherent issues in single \ac{LiDAR} system persist. Section \secref{subsec75} provides an examination of these challenges. Additionally, to enhance robustness, especially in challenging scenarios, researchers have explored the integration of \ac{LiDAR} with other sensor modalities. The integration and its impact on system performance are discussed in more detail in the following Section \secref{sec6}.
\section{Fusion with Other Sensors}\label{sec6}

\ac{LiDAR} demonstrates robustness to changes in lighting conditions, unlike visual sensors; nevertheless, it confronts challenges in demanding environments. Specifically, \ac{LiDAR} odometry encounters difficulties in obtaining accurate measurements under adverse conditions such as rain, snow, and dust. Moreover, \ac{LiDAR} measurements are vulnerable in areas with limited geometric features or repetitive topographical attributes, such as long tunnels or highways. This susceptibility contributes to scan matching challenges, negatively affecting state estimation's precision. Addressing these constraints involves exploring the integration of multiple sensor modalities, marking a notable frontier in current research.

\begin{figure}[!t]
    \centering
    \includegraphics[trim=0 20 0 15, clip, width=1.0\columnwidth]{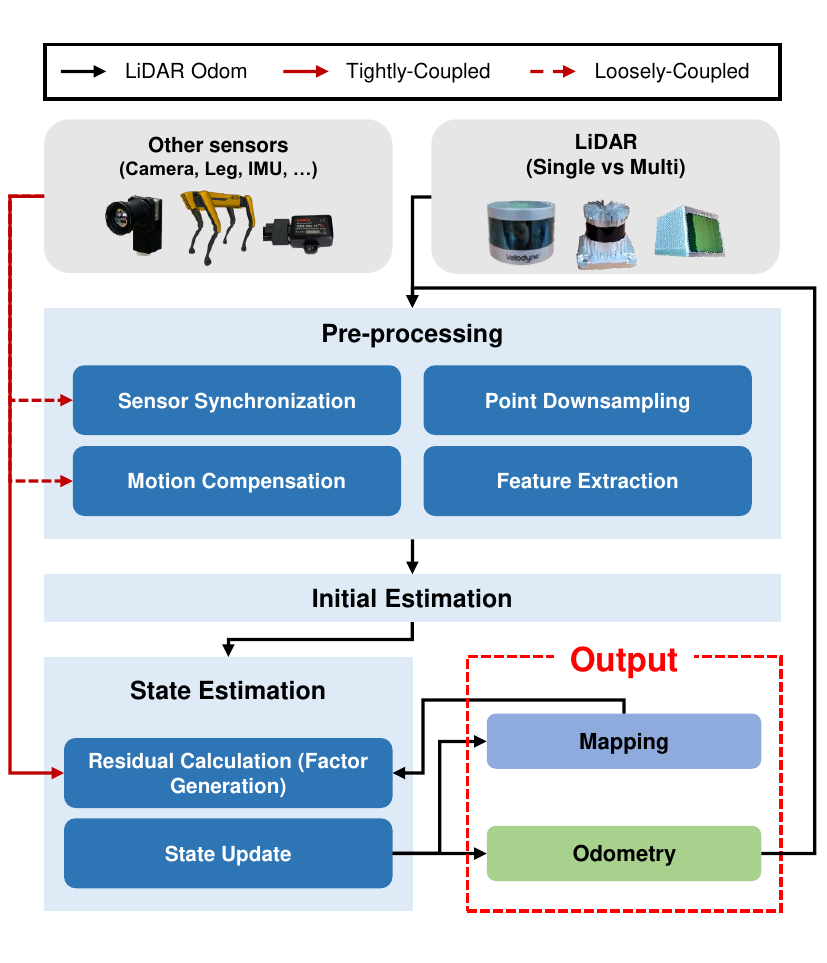}
    \vspace{-1mm}
    \caption{\textbf{LiDAR Odometry Pipeline.} The common framework for LiDAR odometry can be broadly divided into three stages: pre-processing, initial estimation, and state estimation. When incorporating other sensors, their integration is classified as either loosely-coupled or tightly-coupled, based on the specific stage at which the additional sensor data is utilized. In the state estimation stage, the refined state is leveraged both in the odometry and mapping.}
    \label{fig:Categories}
    \vspace{-3mm}
\end{figure}

RGB cameras offer distinct advantages over \ac{LiDAR} sensors, excelling in capturing intricate details through color and texture. This capability becomes crucial in environments where prominent geometric features are scarce. In such scenarios, combining camera images with \ac{LiDAR} measurements can significantly enhance the reliability of state estimation. \citeauthor{lin2021r}. \cite{lin2021r} propose R$^2$live, a tightly-coupled \ac{LiDAR}-visual-inertial odometry system that merges a high-rate filter-based approach with a low-rate graph optimization. The high-rate filter leverages \ac{LiDAR}, camera, and \ac{IMU} measurements, while the factor graph optimizes local maps and visual landmarks. LVI-SAM \cite{lvisam} consists of two jointly operating subsystems: the \ac{LiDAR}-inertial system (LIS) and visual-inertial system (VIS). The estimated pose from each subsystem serves as the initial pose for the other. LIS operates independently only when the number of features in VIS decreases due to aggressive motion or illumination changes, leading to a failure of LIS \cite{7487211}. Similar to R$^2$live, R$^3$live \cite{r3live} also separates the \ac{LiDAR}-inertial odometry (LIO) and visual-inertial odometry (VIO). LIO reconstructs geometric structures, while VIO reconstructs texture information. The proposed VIO system utilizes RGB-colored point cloud maps to estimate the state, minimizing photometric errors without the need to detect visual features, thus saving processing time. Fast-LIVO \cite{fastlivo} enhances efficiency by directly registering point clouds without extracting features. This optimization is achieved by reusing the point clouds from both the LIO and VIO subsystems, resulting in faster operation and improved overall system efficiency. Additionally, LIC-fusion \cite{zuo2019lic, licfusion2} fuses sparse \ac{LiDAR} features with visual features through a multi-state constraint Kalman filter (MSCKF) along with online multi-sensor calibration. In the context of continuous-time \ac{SLAM}, there has been a growing interest in continuous-time \ac{LiDAR}-visual-inertial odometry. An example of such an approach is Coco-LIC \cite{lang2023coco}. This system adopts a non-uniform B-spline-based continuous-time trajectory representation, seamlessly integrating LiDAR and camera data in a tightly-coupled manner.

RGB cameras depend on ambient lighting conditions to capture images, and their performance tends to degrade in low-light or adverse weather conditions. In response to these challenges, thermal cameras operating in the infrared wavelength range have proven effective in visually degraded environments with varying illumination. \citeauthor{rho2023lidar}. \cite{rho2023lidar} utilize stereo thermal cameras in conjunction with \ac{LiDAR} for indoor disaster scenarios. Moreover, radar and event cameras have demonstrated robust performance in challenging environmental conditions. Thermal cameras, radar, and event cameras, when used in conjunction with LiDAR, offer distinct advantages, presenting practical alternatives to address the limitations of RGB cameras. Harnessing these diverse sensor modalities can significantly improve odometry accuracy, as highlighted in \cite{bresson2017simultaneous}.

These sensor modalities extend beyond mobile robots or handheld systems and find application in legged robots. Legged robots excel in navigating bumpy terrains and overcoming obstacles like rocks or debris, leveraging their unique ability to step over them. This capability makes legged robots well-suited for tasks such as search and rescue missions, exploration, and disaster response. VILENS \cite{wisth2022vilens} utilizes measurements from \ac{LiDAR}, IMU, cameras, and leg contact information derived from a joint kinematics model. This integrated sensor fusion empowers the system to attain accurate odometry, even in demanding environments.

Integrating multiple sensors for odometry presents practical solutions for addressing diverse environmental conditions. However, this approach comes with computational demands and introduces specific issues associated with each sensor. While sensor fusion can compensate for the limitations of individual sensors, the fusion process itself requires considerable effort. These limitations will be scrutinized further in Section \secref{subsec75}. Indiscriminate sensor fusion may not lead to an optimal odometry solution. Hence, thorough planning and a precise grasp of each sensor's specific requirements are crucial prerequisites before deploying sensor fusion.

The classifications of LiDAR odometry introduced so far can be organized into a unified pipeline, as shown in the following Figure \ref{fig:Categories}. The figure illustrates how additional sensors can be incorporated into a LiDAR odometry system, guiding the determination of sensor usage and data integration strategies.
\section{Remaining Challenges}\label{sec7}

Undeniably, LiDAR odometry technologies have witnessed significant advancements in providing high-quality positions for mobile robots and autonomous vehicles, with their performance demonstrated in various real-world environments \cite{ebadi2023present, hilti}. However, despite these significant advancements, unresolved issues remain valuable for further research. This section discusses these issues and proposes future directions for LiDAR odometry.

\subsection{LiDAR Inherent Problems}\label{subsec71}
\ac{LiDAR}, while offering accurate measurements and resilience to lighting conditions in contrast to RGB cameras, is not exempt from inherent limitations. In this subsection, we highlight several constraints of the LiDAR sensor that pose challenges in solving the odometry problem.

\textbf{Large Data:} The LiDAR system generates a voluminous 3D point cloud, containing rich environmental and object data. It offers a significant advantage in capturing 3D information about the surrounding environment; however, there are challenges with its size. The size of this point cloud scales with the LiDAR's \ac{FOV} and resolution. For instance, the OS1-128 LiDAR can produce scans containing a substantial number of points, reaching several 100\unit{K} points per frame, operating at a maximum frequency of 20Hz. Additionally, each point in the point cloud includes information such as range, intensity, reflectivity, ambient conditions, and point acquisition time, contributing to the data volume. Real-time processing of such extensive data requires substantial computational power, posing a particular challenge in robotics, where achieving real-time performance is crucial for effective operation.

When integrating multiple LiDARs or adding extra sensors, the computational load is intensified, potentially impacting real-time performance. Techniques such as downsampling or feature extraction can help alleviate the computational burden, but it is evident that computational costs increase with the number and resolution of the LiDARs. In two studies \cite{nguyen2021miliom, nguyen2023slict} utilizing the NTU VIRAL dataset \cite{nguyen2022ntu}, which includes two 16-channel LiDARs, the optimization processes took over 100ms---equivalent to the duration of a LiDAR sweep. While this processing time may be acceptable for systems using keyframes, it becomes impractical in scenarios requiring estimations for every scan.

\textbf{Motion Distortion:} When a robot moves at a high speed relative to the sensor's data acquisition frequency, a substantial spatial gap can occur between the locations where the data was obtained at the beginning and end of a single LiDAR scan. This spatial gap has the potential to introduce significant distortion \cite{al2016analyzing} to the LiDAR scan. Therefore, to effectively utilize LiDAR scans, it is necessary to apply a compensation process to mitigate the distortions caused by motion, commonly referred to as de-skewing. 

De-skewing commonly employs high-frequency sensors such as IMU \cite{shan2020lio, zhang2021lilo} for aligning points to a single frame. Linear interpolation \cite{xu2022fast, nguyen2023slict} can address its discrete nature and the mismatch between sensor measurements and their actual positions. In the absence of extra sensors, a constant velocity model \cite{zhang2014loam, qin2020lins, hong2010vicp} may suffice but lacks accuracy in aggressive motion or uncertain velocity estimations. Continuous-time interpolation \cite{lv2021clins, lang2023coco, zlot2014efficient}, an alternative approach, estimates a continuous trajectory through B-spline interpolation, ensuring accurate transformations for each LiDAR point. However, this method significantly increases computational demands, particularly with more points, as each requires individual state calculation. Thus, balancing accuracy and efficiency is crucial, with the choice depending on the application's specific needs and constraints.

\textbf{Limited Sensing:} LiDAR, while capable of measuring long distances, presents inherent limitations. One prominent drawback is its relatively narrow \ac{FOV}, particularly problematic for perception tasks. Additionally, LiDAR data tends to be sparser than images from standard cameras, even though the horizontal \ac{FOV} is generally wider. 
Recently, advancements in vertical cavity surface emitting laser (VCSEL) technology have enabled the compact arrangement of numerous lasers in a dense array. Despite this advancement, resulting in sensors with increased channels and denser data, the resolution remains lower compared to conventional cameras. In addition, when employing mechanical LiDAR, installations are often in open areas, such as the top of robots or autonomous vehicles, to achieve 360-degree visibility. However, this poses challenges in protecting the sensor from external shocks. Attempts to install the sensor in more sheltered locations result in a trade-off with the loss of \ac{FOV} visibility.

\subsection{Heterogeneous LiDARs}\label{subsec72}
In Section \secref{subsec22}, we discuss the classification of LiDAR sensors into two categories: mechanical and scanning solid-state LiDARs. These categories exhibit distinct characteristics, including variations in viewing angles, scanning patterns, and more. As a result, these disparities essentially lead to the requirement for different odometry algorithms. Moreover, even within the same category of LiDAR, variations in \ac{FOV}, resolutions, and other factors exist across different manufacturers and product lines. This implies that an algorithm effective with one type may necessitate adjustments to additional parameters when applied to another. Recognizing the inconvenience of modifying methods based on the specific sensor, there is a growing demand for an algorithm capable of robust operation across all types of LiDAR.

KISS-ICP \cite{vizzo2023kiss} stands out as a representative approach to addressing these issues. They propose a simplified yet effective LiDAR-only odometry approach that relies on point-to-point ICP, performing comparably with other LiDAR-only methods across various platforms and environmental conditions. Notably, their proposed system is versatile for a broad spectrum of operating conditions using different LiDAR sensors. While KISS-ICP proves to be a simple and versatile solution for various LiDAR sensors, a generalized methodology for LiDAR-inertial odometry and fusing with other sensors is lacking. Consequently, there remains potential for performance improvement in the overall generalized approaches.

\subsection{Degenerative Environment}\label{subsec73}
Traditional LiDAR odometry primarily depends on geometric measurements, neglecting texture and color information usage. This reliance becomes challenging in feature-scarce and repetitive environments, such as tunnels and long corridors. While LiDAR effectively performs scanning in these settings, the absence of unique features often leads to ambiguity in scan matching, resulting in potential inaccuracies in the pose estimation of robots.

To tackle this challenge, \citeauthor{7487211}. \cite{7487211} introduce a mathematical definition of degeneracy factor derived and evaluated using eigenvalues and eigenvectors, enabling more accurate state estimation when a degeneracy is detected. AdaLIO \cite{lim2023adalio} introduces an adaptive parameter setting strategy, advocating for the use of environment-specific parameters to address the degeneracy issue. Their straightforward approach involves pre-defining parameters for general and degenerate scenarios and adjusting them based on the situation. \citeauthor{wang2023hierarchical}. \cite{wang2023hierarchical} mitigate the uncertainty associated with the corresponding residual and address the degeneration problem by removing eigenvalue elements from the distribution covariance component. \citeauthor{9996434}. \cite{9996434} propose an adaptive correlative scan matching (CSM) algorithm that dynamically adjusts motion weights based on degeneration descriptors, enabling autonomous adaptation to different environments. This approach aligns the initial pose weight with environmental characteristics, resulting in improved odometry results.

Sensor fusion methods also have shown the potential to address the uncertainty in LiDAR scan matching within degenerative cases. DAMS-LIO \cite{han2023dams} estimates LiDAR-inertial odometry utilizing the \ac{iEKF}. When the system detects degeneration, it employs a sensor fusion strategy, following a loosely-coupled approach that integrates odometry results from each sensor. 

LiDAR has the potential to overcome degenerative environments without the need for sensor fusion if additional information can be accessed from the measurements beyond the geometric details. Researchers have explored leveraging intensity \cite{wang2021intensity, park2020loam, li2022intensity} or reflectivity \cite{dong2023r, zhang2023ri} data from LiDAR measurements to enhance state estimation in degenerate environments. Integrating supplementary texture information with the original geometric data offers a more robust and reliable solution, particularly in challenging scenarios where geometric features alone may not suffice for accurate localization and mapping. Furthermore, by employing FMCW LiDAR to measure Doppler velocity similar to radar, DICP \cite{hexsel2022dicp} improves the vanilla ICP algorithm with a Doppler velocity objective term, enhancing scan matching performance, especially in feature-scarce environments. Notably, their work forecasts odometry with high accuracy, even in the demanding scenario of a 900-meter-long tunnel sequence. Improving upon DICP, \citeauthor{wu2022picking}. \cite{wu2022picking}. and \citeauthor{yoon2023need}. \cite{yoon2023need} integrate the Doppler velocity factor in a continuous-time odometry framework. These works suggest that the degeneracy problem can be effectively addressed through the use of FMCW LiDAR.

\subsection{Degraded Environment}\label{subsec74}
A degraded environment is one that presents challenges to the sensing ability of LiDAR, unlike a degenerative environment. LiDAR operates by emitting a laser pulse and detecting its return after interacting with objects, and this process can be disrupted by unwanted particles obstructing the pulse's path. Extreme weather conditions such as direct sunlight, rain, snow, or fog can significantly degrade LiDAR's detection performance \cite{sun2016technique, bijelic2018benchmark}. Considerable research has been dedicated to denoising weather-induced interferences to address this challenge due to extreme weather. \citeauthor{park2020fast}. \cite{park2020fast} propose a Low-Intensity Outlier Removal (LIOR) filter to eliminate snow particles from the LiDAR point cloud. Utilizing a CNN-based approach, WeatherNet \cite{heinzler2020cnn}, a variant of LiLaNet \cite{piewak2018boosting}, is trained with augmented data incorporating a fog model and a rain model. This training process aims to effectively remove noise caused by adverse weather conditions from the actual LiDAR data. Despite extensive research on weather noise removal algorithms, there is a lack of investigation into the performance of LiDAR odometry using these algorithms. Exploring this area is essential to ensure that LiDAR odometry consistently delivers high-level performance under harsh weather conditions, ensuring the stability of autonomous driving.

Beyond weather conditions, typical objects, such as glass, that partially reflect or transmit laser pulses \cite{zhao2020mapping, weerakoon2022cartographer_glass, foster2013visagge} can adversely affect LiDAR performance. This problem is particularly prevalent in urban or indoor settings with numerous glass windows, where reflections from one side can interfere with the LiDAR points on the opposite side of the glass. This issue can impact odometry performance due to the ambiguity in scan matching. However, there is currently a lack of research on algorithms to address this problem completely.

\subsection{Multi-Modal Sensors}\label{subsec75}
When integrating additional sensors with LiDAR, it is crucial to acknowledge that these supplementary sensors introduce their own set of challenges. Moreover, the combination of multiple sensors can introduce new limitations and complexities. This subsection delves into these additional considerations.

\textbf{Calibration:} When working with multiple sensors, it is essential to conduct both intrinsic calibration for each sensor and extrinsic calibration between the sensors. However, it is crucial to note that this calibration process can be highly challenging and complex despite the availability of calibration tools and methodologies \cite{9779777, rehder2016extending, domhof2019extrinsic}. Precise intrinsic calibration for each sensor and accurate extrinsic calibration between multiple sensors present difficulties involving addressing diverse error sources, considering environmental factors, and managing complex mathematical transformations. The intricacies of calibration can make the process time-consuming and demanding for both researchers. Even with precise calibration tools, calibrating sensors in systems where the system itself cannot impose constraints on each sensor can be problematic. For instance, car-like vehicles often have insufficient constraints for the z-axis, roll, and pitch angles. As a result, the accuracy of these elements may not surpass that achieved through manual measurements.

\textbf{Placement:} Simply adding more sensors without strategic planning may not affect odometry performance. In the case of multiple LiDARs, strategic positioning to complement scanning areas has the potential for accuracy improvement. However, excessive overlap can lead to redundancy, introducing unnecessary data and increasing computational costs, potentially offsetting accuracy gains \cite{jung2023asynchronous}. Therefore, careful consideration of optimal deployment strategies is crucial. Although \citeauthor{hu2022investigating}. \cite{hu2022investigating} discuss effective multi-LiDAR placement strategies; their focus is on object detection rather than odometry research. Hence, dedicated studies in this domain are needed. This challenge also extends to multi-modal sensor fusion. Similar to the placement considerations for multiple LiDARs, the configuration of each sensor is crucial in system design. Different sensors serve unique roles with diverse recognition capabilities. To maximize the strengths of different sensors, careful consideration is essential in determining whether each sensor's \ac{FOV} should overlap.

\textbf{Synchronization:} Integrating different sensor modalities necessitates addressing asynchronous scenarios, as each sensor delivers data in distinct frequencies. While some studies adeptly fuse heterogeneous LiDAR data in discrete-time \cite{nguyen2021miliom} or continuous-time \cite{jung2023asynchronous} using IMU, there is a relatively limited body of work on the integration of various sensor modalities. Exploring comprehensive approaches to harness the capabilities of different sensor modalities holds significant potential.
\section{Datasets \& Evaluation}\label{sec8}

\begin{table*}[!t]
\centering
\caption{The LiDAR-based odometry-related datasets. \textbf{S}, \textbf{US} and \textbf{IN} represents the structured, unstructured and indoor environments. Furthermore, the number of $\filledstar$ varies depending on whether it is under 10km, between 10 and 100km, and over 100km.}
\label{tab:dataset}
\resizebox{\textwidth}{!}{%
\begin{tabular}{ccccccc} 
\toprule
\multicolumn{1}{c|}{\multirow{2}{*}{Dataset}} &
  \multicolumn{1}{c|}{\multirow{2}{*}{System}} &
  \multicolumn{1}{c|}{\multirow{2}{*}{Scenario}} &
  \multicolumn{2}{c|}{Sensor Input (LiDAR)} &
  \multicolumn{1}{c|}{\multirow{2}{*}{Groundtruth}} &
  \multirow{2}{*}{Distance} \\ 
\multicolumn{1}{c|}{} &
  \multicolumn{1}{c|}{} &
  \multicolumn{1}{c|}{} &
  \# Spinning &
  \multicolumn{1}{c|}{\# Solid State} &
  \multicolumn{1}{c|}{} &
   \\ \midrule
\multicolumn{1}{c|}{KITTI \cite{kitti}} &
  \multicolumn{1}{c|}{Car} &
  \multicolumn{1}{c|}{\textbf{S}} &
  1 &
  \multicolumn{1}{c|}{-} &
  \multicolumn{1}{c|}{GPS/IMU} &
  $\filledstar$ $\filledstar$ \\
\multicolumn{1}{c|}{NCLT \cite{nclt}} &
  \multicolumn{1}{c|}{Segway} &
  \multicolumn{1}{c|}{\textbf{S}} &
  1 &
  \multicolumn{1}{c|}{-} &
  \multicolumn{1}{c|}{RTK-GPS/ICP} &
  $\filledstar$ $\filledstar$ $\filledstar$ \\
\multicolumn{1}{c|}{Complex Urban \cite{complex}} &
  \multicolumn{1}{c|}{Car} &
  \multicolumn{1}{c|}{\textbf{S}} &
  2 &
  \multicolumn{1}{c|}{-} &
  \multicolumn{1}{c|}{VRS-GPS/FOG/Encoder/ICP} &
  $\filledstar$ $\filledstar$ $\filledstar$ \\
\multicolumn{1}{c|}{MulRan \cite{mulran}} &
  \multicolumn{1}{c|}{Car} &
  \multicolumn{1}{c|}{\textbf{S}} &
  1 &
  \multicolumn{1}{c|}{-} &
  \multicolumn{1}{c|}{VRS-GPS/FOG/ICP} &
  $\filledstar$ $\filledstar$ $\filledstar$ \\
\multicolumn{1}{c|}{Oxford Radar Robotcar \cite{oxford}} &
  \multicolumn{1}{c|}{Car} &
  \multicolumn{1}{c|}{\textbf{S}} &
  2 &
  \multicolumn{1}{c|}{-} &
  \multicolumn{1}{c|}{GPS/INS/SLAM} &
  $\filledstar$ $\filledstar$ $\filledstar$ \\
\multicolumn{1}{c|}{Ford AV \cite{agarwal2020ford}} &
  \multicolumn{1}{c|}{Car} &
  \multicolumn{1}{c|}{\textbf{S} + \textbf{US}} &
  4 &
  \multicolumn{1}{c|}{-} &
  \multicolumn{1}{c|}{GPS/IMU/SLAM} &
  $\filledstar$ $\filledstar$ $\filledstar$ \\
\multicolumn{1}{c|}{LIBRE \cite{libre}} &
  \multicolumn{1}{c|}{Car} &
  \multicolumn{1}{c|}{\textbf{S}} &
  12 &
  \multicolumn{1}{c|}{-} &
  \multicolumn{1}{c|}{GPS} &
  $\filledstar$ $\filledstar$ $\filledstar$ \\
\multicolumn{1}{c|}{EU Longterm \cite{eulongterm}} &
  \multicolumn{1}{c|}{Car} &
  \multicolumn{1}{c|}{\textbf{S}} &
  2 &
  \multicolumn{1}{c|}{-} &
  \multicolumn{1}{c|}{RTK-GPS/IMU} &
  $\filledstar$ $\filledstar$ \\
\multicolumn{1}{c|}{NTU Viral \cite{nguyen2022ntu}} &
  \multicolumn{1}{c|}{Drone} &
  \multicolumn{1}{c|}{\textbf{S} + \textbf{IN}} &
  2 &
  \multicolumn{1}{c|}{-} &
  \multicolumn{1}{c|}{Tracker} &
  $\filledstar$ \\
\multicolumn{1}{c|}{UrbanNav \cite{urbannav}} &
  \multicolumn{1}{c|}{Car} &
  \multicolumn{1}{c|}{\textbf{S}} &
  3 &
  \multicolumn{1}{c|}{-} &
  \multicolumn{1}{c|}{RTK-GPS/INS} &
  $\filledstar$ $\filledstar$ \\
\multicolumn{1}{c|}{Hilti-Oxford \cite{hilti}} &
  \multicolumn{1}{c|}{Handheld} &
  \multicolumn{1}{c|}{\textbf{S} + \textbf{IN}} &
  1 &
  \multicolumn{1}{c|}{-} &
  \multicolumn{1}{c|}{TLS/GCP} &
  $\filledstar$ \\
\multicolumn{1}{c|}{Tiers \cite{tiers}} &
  \multicolumn{1}{c|}{Car} &
  \multicolumn{1}{c|}{\textbf{S} + \textbf{US} + \textbf{IN}} &
  3 &
  \multicolumn{1}{c|}{3} &
  \multicolumn{1}{c|}{Tracker/SLAM} &
  $\filledstar$ \\
\multicolumn{1}{c|}{Wild Places \cite{wild}} &
  \multicolumn{1}{c|}{Handheld} &
  \multicolumn{1}{c|}{\textbf{US}} &
  1 &
  \multicolumn{1}{c|}{-} &
  \multicolumn{1}{c|}{SLAM} &
  $\filledstar$ $\filledstar$ \\
\multicolumn{1}{c|}{Pohang Canal \cite{pohang}} &
  \multicolumn{1}{c|}{Ship} &
  \multicolumn{1}{c|}{\textbf{S}} &
  3 &
  \multicolumn{1}{c|}{-} &
  \multicolumn{1}{c|}{RTK-GPS/IMU} &
  $\filledstar$ $\filledstar$ \\
\multicolumn{1}{c|}{ConSLAM \cite{trzeciak2022conslam}} &
  \multicolumn{1}{c|}{Handheld} &
  \multicolumn{1}{c|}{\textbf{S + IN}} &
  1 &
  \multicolumn{1}{c|}{-} &
  \multicolumn{1}{c|}{TLS} &
  $\filledstar$ \\
\multicolumn{1}{c|}{Boreas \cite{boreas}} &
  \multicolumn{1}{c|}{Car} &
  \multicolumn{1}{c|}{\textbf{S}} &
  1 &
  \multicolumn{1}{c|}{-} &
  \multicolumn{1}{c|}{RTK-GPS/INS} &
  $\filledstar$ $\filledstar$ $\filledstar$ \\
\multicolumn{1}{c|}{HeLiPR \cite{helipr}} &
  \multicolumn{1}{c|}{Car} &
  \multicolumn{1}{c|}{\textbf{S}} &
  2 &
  \multicolumn{1}{c|}{2} &
  \multicolumn{1}{c|}{RTK-GPS/INS} &
  $\filledstar$ $\filledstar$ $\filledstar$ \\ \bottomrule
\end{tabular}%
}
\vspace{-2mm}
\end{table*}

Ensuring the generalization of LiDAR odometry remains a fundamental goal in its advancement, as elaborated in Section \secref{sec7}. As autonomous systems navigate diverse and dynamic environments, algorithms must exhibit consistent performance, irrespective of variations in data quality. Consequently, the significance of comprehensive datasets spanning various environments and sensor modalities cannot be overstated in the development of such algorithms. Diverse data enhances robustness, reducing the risk of overfitting and expanding the versatility of techniques. Simultaneously, establishing standardized evaluation methodologies is crucial to ensure consistent and comparable results across diverse research endeavors. With the growing role of LiDAR odometry in robotics, there is an increased emphasis on creating diverse datasets and refining assessment protocols. These strategic initiatives are essential for effectively addressing various operational challenges.

\subsection{Public Datasets}\label{subsec81}
Various LiDAR datasets have contributed significantly to odometry research, each with unique features and limitations. In this section, we will present public LiDAR datasets along with their respective characteristics. Public LiDAR datasets are summarized in \tabref{tab:dataset}.

The KITTI dataset \cite{kitti}, which captures the urban environments using the HDL-64E spinning LiDAR, stands as a renowned resource in the LiDAR community. Its overlapping sequences within and between sessions facilitate precise odometry evaluation, contributing significantly to LiDAR odometry advancements. The NCLT dataset \cite{nclt}, collected over a year using a Segway-based system, and the MulRan dataset \cite{mulran}, spanning around a month, offer spatial diversity from campuses to cityscapes. The Boreas dataset \cite{boreas}, collected over a year on cityscape routes, captures seasonal changes and harsh weather conditions such as rain and heavy snow. While they are invaluable resources for developing odometry algorithms tailored to a specific LiDAR type, these datasets are limited in terms of LiDAR hardware diversity, predominantly relying on a single type of LiDAR. This poses a challenge for algorithms aiming to achieve broader hardware compatibility.

The Complex Urban dataset \cite{complex}, Oxford Radar Robotcar dataset \cite{oxford}, and EU Long-term dataset \cite{eulongterm} distinguish themselves from previous datasets as they utilize multiple LiDARs. The Complex Urban dataset captures data across various urban environments, while the Oxford Radar Robotcar dataset focuses on data collection from a single location for consistency. The EU Long-term dataset, spanning data acquisition over two locations for approximately a year, showcases diverse weather conditions. Despite using multiple LiDARs in these datasets, challenges in generalization arise due to their consistent use of homogeneous LiDAR configurations and a focus on structured environments. This raises concerns about the performance of LiDAR odometry in diverse settings. 

The Ford AV dataset \cite{agarwal2020ford} addresses the previously mentioned limitation by ensuring location diversity. It captures seasonal variations and various driving scenarios, encompassing freeways, residential areas, tunnels, and vegetation-rich zones, utilizing four HDL-32E LiDARs. Nevertheless, the uniform configuration of the LiDARs still poses a challenge. In contrast, LIBRE \cite{libre} provides a driving dataset along with a separate distance error report for 12 LiDARs, detailing performance under diverse weather conditions. It is essential to note that each sequence in LIBRE features only a single LiDAR. Moreover, the dataset does not provide insights into LiDAR odometry on platforms with aggressive motions since it only involves stationary LiDAR-equipped vehicles in artificially controlled weather conditions.

The previously mentioned LiDAR datasets collected with mapping-car systems have limitations in roll and pitch angle variations. To address this, the NTU Viral dataset \cite{nguyen2022ntu} introduces a new challenge by deploying LiDAR on an unmanned aerial vehicle (UAV). Similarly, the Hilti-Oxford dataset \cite{hilti}, ConSLAM dataset \cite{trzeciak2022conslam}, and Wild Places dataset \cite{wild} present this challenge using a handheld system in construction site and forest. The Hilti-Oxford dataset further diversifies the landscape by including data from indoor environments, while Wild Places ventures into forest terrains, adding complexity to the dataset landscape. Additionally, the Pohang Canal dataset \cite{pohang} captures canal environments using a ship-based system.

Recent LiDAR datasets have introduced new dimensions to research by incorporating multiple heterogeneous LiDARs. For instance, the UrbanNav dataset \cite{urbannav} features three mechanical LiDARs navigating urban landscapes, presenting challenges due to asynchronous multiple LiDARs. The Tiers dataset \cite{tiers} employs a combination of three mechanical and three scanning solid-state LiDARs, capturing distinct measurements from identical locations and offering a unique perspective. On a larger scale, the HeLiPR dataset \cite{helipr} includes a variety of structured environments and introduces FMCW LiDAR, providing the opportunity to utilize velocity information for LiDAR odometry.

Various LiDAR odometry datasets with unique strengths and limitations have been released and continue to emerge. This highlights the importance of recognizing that no single dataset can offer universal comprehensiveness. Thus, the thoughtful selection of datasets aligned with their specific characteristics remains essential for the development of robust and adaptable LiDAR odometry solutions. 

\subsection{Evaluation}\label{subsec82}
Evaluation serves as a cornerstone in advancing LiDAR odometry. Comprehensive and consistent evaluation methods are essential, as they enable the measurement of progress, identification of weaknesses, and guidance for future research. Assessing LiDAR odometry algorithms is crucial for establishing their dependability and accuracy and fostering comparability across various approaches. Ultimately, this promotes continuous improvement in LiDAR odometry.

\begin{figure*}[!t]
    \centering
    \subfloat[Trajectory Error\label{fig:Traj}]{
		\includegraphics[width=0.5\columnwidth]{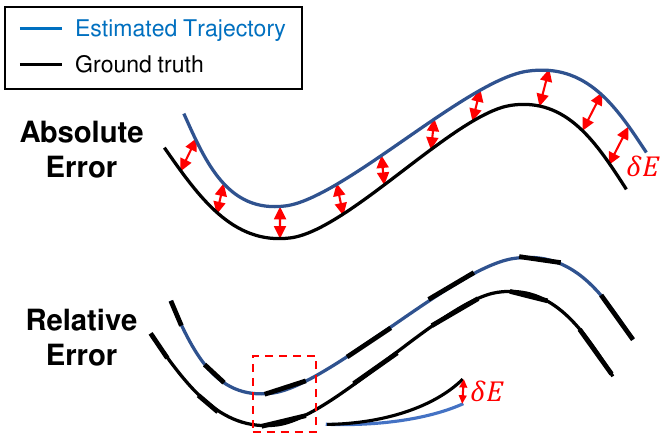}
    }
    \subfloat[Start-to-End Error\label{fig:Start}]{
		\includegraphics[width=0.5\columnwidth]{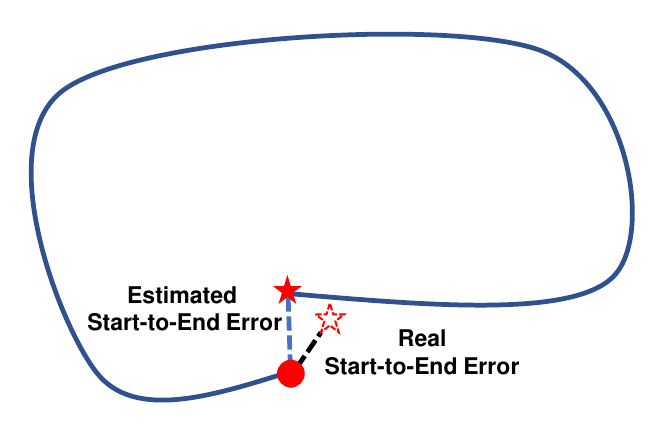}
	}
    \subfloat[GCP-based Error\label{fig:GCP}]{
		\includegraphics[width=0.5\columnwidth]{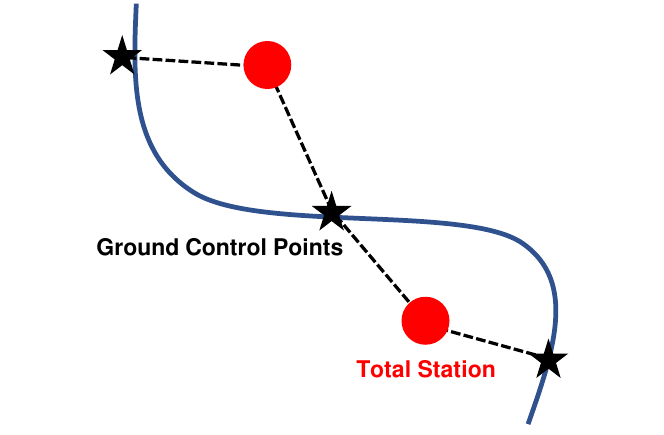}
	}
 \subfloat[Entrophy-based Error \label{fig:Enthropy}]{
		\includegraphics[trim={0 1cm 0 1cm},clip, width=0.5\columnwidth]{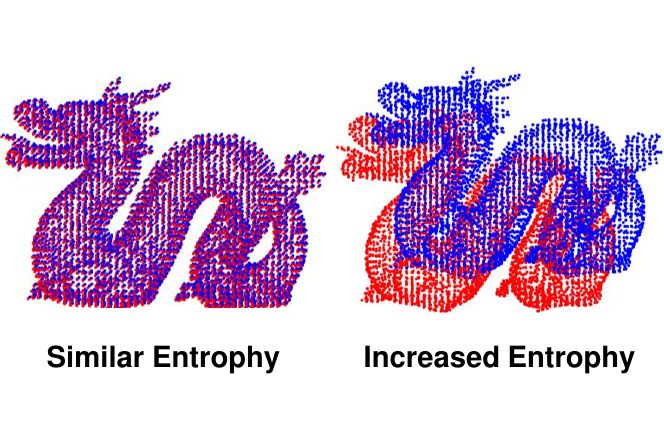}
	}

    \vspace{-1mm}
    \caption{\textbf{Various Evaluation Methods of LiDAR Odometry.} This figure illustrates diverse assessment methods for LiDAR odometry. (a) Trajectory Error shows local and global discrepancies along the estimated path. (b) Start-to-End Error highlights long-term drifts from start to finish. (c) GCP-based Error assesses alignment with GCPs for real-world accuracy. (d) Entropy-based Error reflects scan registration quality and overall system reliability.}
    \label{fig:feature}
    \vspace{-5mm}
\end{figure*}

\subsubsection{Groundtruth generation}
The cornerstone of the evaluation process is the ground truth, serving as the reference to assess the precision and reliability of odometry estimation. Various methods can be employed to reliably evaluate LiDAR odometry to obtain ground truth, each with unique strengths and potential limitations.

One approach utilizes \ac{GPS}, providing precise global position measurements. When combined with \ac{RTK} compensation, \ac{GPS} can attain centimeter-level precision. Integration of \ac{GPS} with an \ac{IMU} enables the derivation of a complete 6-\ac{DOF} pose, encompassing both position and orientation. Additionally, the integration of \ac{INS} further improves continuous pose estimation, particularly in environments with weak or lost \ac{GPS} signals.

Another method involves leveraging \ac{SLAM} technology. The trajectory generated by \ac{SLAM}, utilizing sensors such as LiDAR, camera, and encoder, can serve as an additional reference for ground truth, especially in environments where \ac{GPS} signals are unavailable. Combining the strengths of both \ac{GPS} and \ac{SLAM} can create a robust system that offers high accuracy and resilience to environmental challenges.

A third approach entails employing tracking systems. These specialized systems, typically optical, utilize multiple cameras \cite{tiers} or sensors \cite{nguyen2022ntu} to meticulously track markers or objects within a designated area. They prove especially valuable in environments with low \ac{SLAM} accuracy or where \ac{GPS} signals are unavailable. Due to their firmly established precision in both temporal and spatial dimensions, tracking systems become a reliable reference for ground truth in controlled setups.

The Ground Control Points (GCP) method constitutes the fourth approach. This method utilizes specific ground points with known and precise geographical locations, often established using total stations. These GCPs are frequently employed to guarantee accurate positioning and alignment. By comparing sensor data with these reference points, any discrepancies can be identified and corrected, ensuring high measurement accuracy.

Finally, Terrestrial Laser Scanning (TLS) is utilized to establish ground truth. As a variant of LiDAR, TLS swiftly scans and captures 3D data of the environment. Due to its extensive reach and high-resolution data, TLS-based ground truth serves as a benchmark for aligning individual scans. The alignment of these scans to the TLS-derived ground truth enables the determination of the robot's 6-DOF state, which then serves as the definitive reference for LiDAR odometry.

\subsubsection{Evaluation Methods}
In the evaluation of LiDAR odometry, several quantitative metrics are pivotal for assessing the accuracy and effectiveness of algorithms. When compared to reliable ground truth, these metrics offer insights into the precision, stability, and areas for potential improvement of a particular odometry system. This section will explore some essential evaluation methods.

Initially, we consider the \ac{ATE}. \ac{ATE} provides a comprehensive perspective on the overall odometry consistency. It computes the average deviation between corresponding poses in the estimated trajectory relative to the ground truth, thereby capturing discrepancies throughout the trajectory. Mathematically, it is expressed as:
\begin{equation}
ATE = \sqrt{\frac{1}{N} \sum_{i=1}^{N} || p_{i, \text{est}} - p_{i, \text{gt}} ||^2}
\label{eq:ate}
\end{equation}
where \( p_{i, \text{est}} \) represents the estimated pose, \( p_{i, \text{gt}} \) the ground truth pose, and \( N \) the total number of poses.

Next, our focus shifts to the \ac{RTE}. Unlike the broad scope of \ac{ATE}, \ac{RTE} concentrates on shorter segments of the trajectory. It evaluates the local consistency and accuracy of the odometry, which is particularly crucial for applications that require precision over shorter distances. The formulation of \ac{RTE} can be represented as:
\begin{equation}
RTE = \sqrt{\frac{1}{M} \sum_{j=1}^{M} || q_{j, \text{est}} - q_{j, \text{gt}} ||^2}
\label{eq:rte}
\end{equation}
where \( q_{j, \text{est}} \) and \( q_{j, \text{gt}} \) respectively denote estimated and ground truth relative poses over a defined segment, with \( M \) being the number of such segments. The \ac{ATE} and \ac{RTE} are typically calculated using the RPG \cite{Zhang18iros} and EVO evaluator \cite{grupp2017evo}.

The Start-to-End Error proves particularly insightful in assessing the long-term consistency and reliability of the odometry result. This metric evaluates the misalignment between the initial and final points of trajectories, offering a macroscopic perspective on odometry performance. Notably, as precisely locating the exact start and end points can be challenging, the error is determined by computing the relative translation between these points using registration methods such as ICP and Generalized-ICP. It is formulated as:
\begin{equation}
\text{Error} = \left| \Delta \mathbf{p}_{\text{est}} - \Delta \mathbf{p}_{\text{gt}} \right|
\end{equation}
where \(\Delta \mathbf{p}_{\text{est}}\) and \(\Delta \mathbf{p}_{\text{gt}}\) are the position differences calculated from odometry and registration method. It is particularly effective when facing challenges in obtaining a reliable ground truth for the trajectory. Such challenges may arise in indoor environments where obtaining GPS measurements is problematic or in unstructured terrains where the accuracy of SLAM is compromised.

Another approach utilizes GCPs, predetermined precise ground locations typically established with total stations. To conduct an evaluation using GCPs, the estimated trajectory undergoes alignment with these control points using SE(3) Umeyama alignment \cite{umeyama}. Following alignment, the absolute distance error for each GCP is calculated to gauge its deviation from the predicted trajectory. This method hinges on the precision of GCPs to assess the accuracy of the odometry system.

Lastly, certain methods assess the registration quality between consecutive scans. Given that the trajectory derived from LiDAR odometry depends on successful registration, evaluating this aspect can indirectly provide insights into odometry accuracy. The concept of entropy \cite{corAl} serves as a valuable tool for such evaluations. When two point clouds are accurately registered, the merged point cloud retains entropy similar to that of the original individual point clouds. In contrast, poor registration leads to higher entropy in the combined point cloud. This demonstrates that appropriately registered point clouds maintain consistent entropy, or uncertainty, in their combined form, making it a valuable metric for evaluating registration quality.

Each evaluation method for LiDAR odometry offers distinct insights. Researchers must choose the most suitable validation approach based on their specific experimental context. Continuous advancements and the introduction of innovative comparison methodologies have the potential to enhance the comprehensive evaluation of robustness and accuracy over time.

\begin{table}[t]
\vspace{-5mm}
\centering
\caption{Benchmark Results (up: LO, down: LIO)}
\label{tab:benchmark}
\begin{tabular}{c|ccc} \toprule
\multirow{2}{*}{\textbf{Algorithms}} & \multicolumn{3}{c}{Evaluation Metric: ATE (m)}                                                       \\
                                                                       &ConSLAM         & NTU VIRAL         & HeLiPR \\ 
 \midrule
 LOAM                                 &      -          &        0.959      &     23.043       \\
                                  LeGO-LOAM                            &       0.263         &       8.478       &     10.111        \\
                                  KISS-ICP                             &       13.517         &      0.829       &     29.273        \\
                                  DLO                                  &        0.154       &    0.142       &      5.022        \\
                                  VoxelMap                             &       -         &       1.100         &    5.010          \\ \midrule

                                  LIO-SAM                              &       0.167        &       0.115         &    42.414         \\
                                  FAST-LIO2                            &       0.113         &    0.116        &    1.655         \\
                                  Faster-LIO                           &       0.102        &     0.120            &    22.402         \\
                                  DLIO                                 &       0.106         &    0.224           &     2.042      \\
                                  Point-LIO                            &       0.115        &    0.105             &      17.142      \\ \bottomrule
\end{tabular}%

\label{tab:benchmark}
\vspace{-2mm}

\end{table}

\begin{figure}[!t]
    \centering
    \subfloat[Trajectories with HeLiPR dataset \label{fig:helipr}]{
		\includegraphics[trim=15 15 0 20, clip, width=1.0\columnwidth]{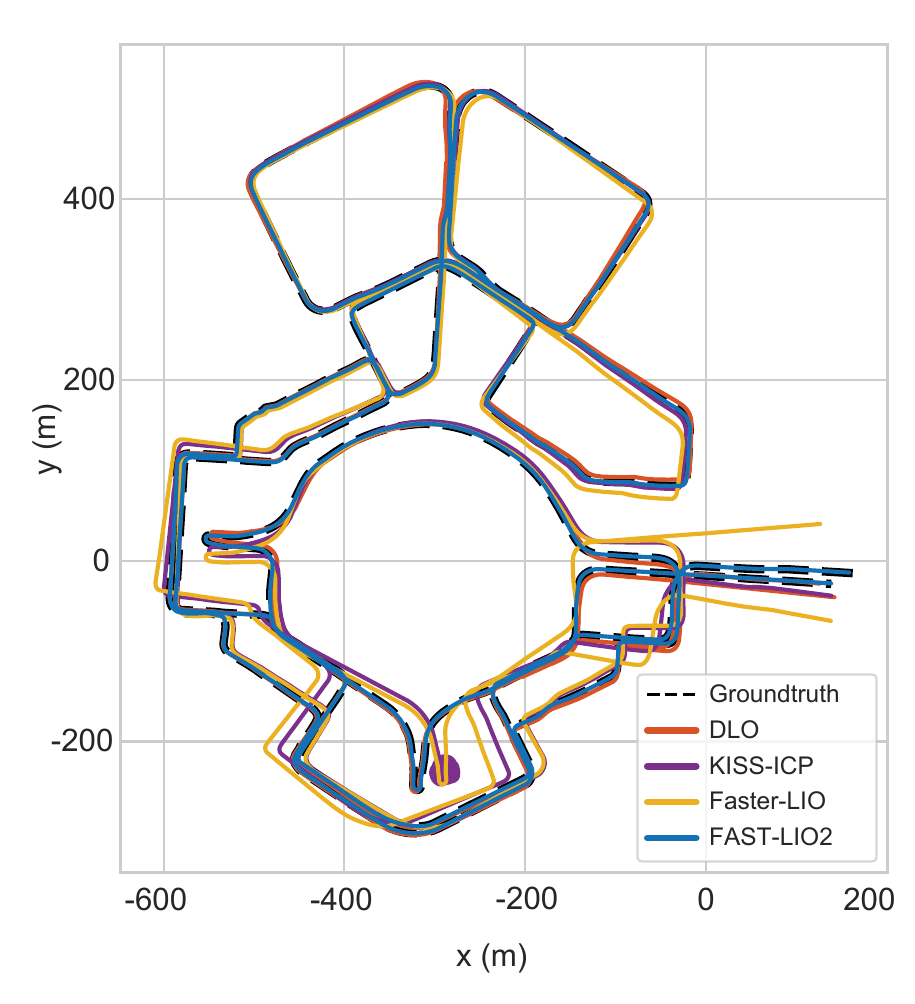}
    }\\
    \subfloat[Trajectories with ConSLAM dataset\label{fig:conslam}]{
		\includegraphics[trim=20 20 5 50, clip, width=1.0\columnwidth]{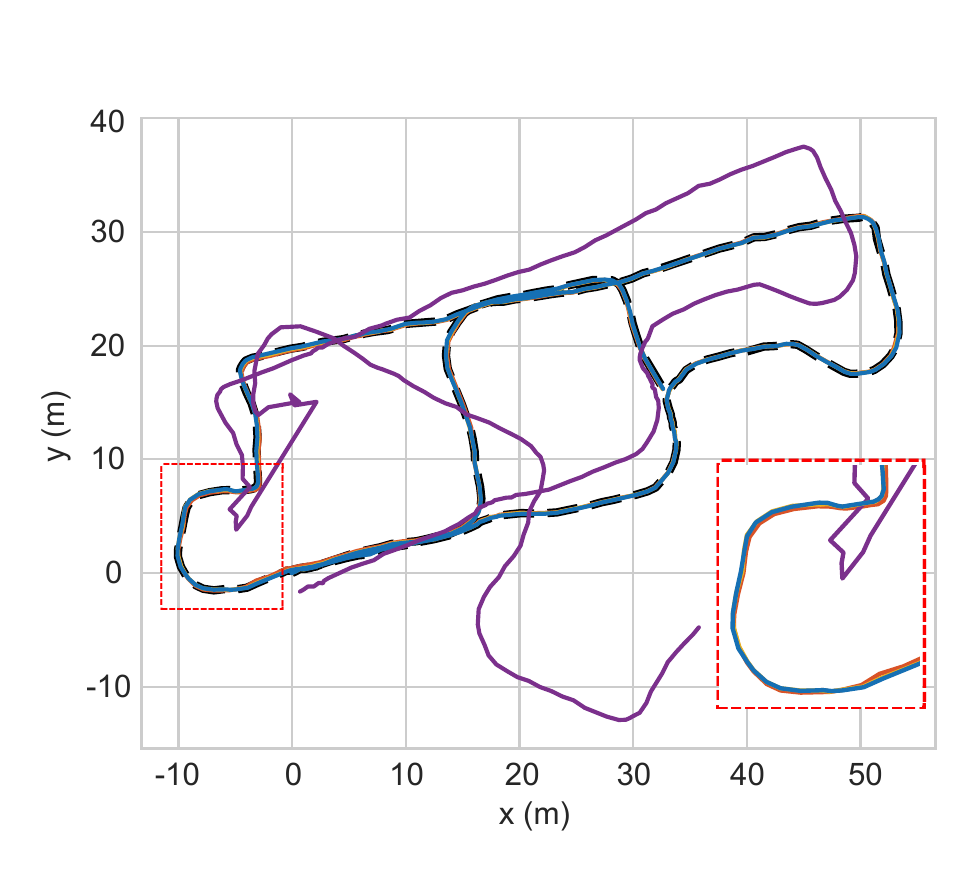}
    }
    \vspace{-1mm}
    \caption{\textbf{Estimated Trajectories.} This figure illustrates results from the HeLiPR and ConSLAM datasets. In Figure (b), the red box zooms in on instances where the LiDAR odometry deviates from the ground truth compared to LiDAR-inertial odometry.}
    \label{fig:benchmark}
    \vspace{-5mm}
\end{figure}

\subsection{Benchmark Results}
Based on the aforementioned datasets and evaluation methods, we conduct benchmarks to compare the performance of LiDAR-only odometry and LiDAR-inertial odometry. Our benchmark test analyzes the performance of six LiDAR-only odometry and six LiDAR-inertial odometry literature. For LiDAR-only odometry, the selected methods are LOAM \cite{zhang2014loam}, LeGO-LOAM \cite{shan2018lego}, KISS-ICP \cite{vizzo2023kiss}, CT-ICP \cite{dellenbach2022ct} and DLO \cite{9681177}. In the case of LiDAR-inertial odometry, our focus is on LIO-SAM \cite{shan2020lio}, FAST-LIO2 \cite{xu2022fast}, VoxelMap \cite{yuan2022efficient}, DLIO \cite{chen2023direct}, and Point-LIO \cite{he2023point}. 

The evaluation of LiDAR-only and LiDAR-inertial odometry works, as shown in \tabref{tab:benchmark}, has been performed on \texttt{sequence02} from the ConSLAM dataset \cite{trzeciak2022conslam}, \texttt{eee03} from the NTU VIRAL dataset \cite{nguyen2022ntu}, and \texttt{Roundabout02} from the HeLiPR dataset \cite{helipr}. We select these three datasets for evaluation due to their distinct characteristics. The ConSLAM dataset captures brief sequences from a construction site using a handheld system, the NTU VIRAL dataset acquires short sequences from a campus via a drone, and the HeLiPR dataset utilizes a car for large-scale data acquisition at the city level. It is essential to emphasize the variations in both systems and environments used for data acquisition across these datasets.

We assess method performance by measuring the ATE in meters. The NTU VIRAL dataset employs its dedicated evaluation tool for measurements, while for other datasets, we use EVO \cite{grupp2017evo}, a widely recognized tool in the field.

As evidenced in \tabref{tab:benchmark}, LiDAR-inertial odometry generally demonstrates enhanced robustness compared to LiDAR-only odometry. However, it is important to note that not all LiDAR-inertial systems outperform LiDAR-only systems, particularly within the HeLiPR dataset. The sequence from the HeLiPR dataset, being exceptionally long, is susceptible to cumulative errors as indicated in \figref{fig:helipr}. In such cases, integrating an IMU with LiDAR may not significantly outperform LiDAR-only odometry due to potential error accumulation after large drift. This highlights the necessity of integrating error-resolving mechanisms such as GPS or loop closure in prolonged robot operations to improve odometry performance.

On the other hand, fusion with an IMU can enhance accuracy for shorter paths. The notable advantage of LiDAR-inertial odometry lies in its effective handling of aggressive motions, especially sudden rotations. This becomes particularly evident in scenarios with dynamic motion, such as those involving handheld systems or drones in datasets like ConSLAM or NTU VIRAL. In ConSLAM, although some LiDAR-only and LiDAR-inertial methods may exhibit similar paths, a closer examination reveals that LiDAR-only odometry lacks precision in detailed path estimation. It deviates more significantly from the ground truth compared to LiDAR-inertial odometry, as depicted in \figref{fig:conslam}.

In summary, while LiDAR-inertial odometry generally surpasses LiDAR-only systems in robustness, it does not correctly estimate in all scenarios, especially in long sequences prone to cumulative errors. In contrast, for shorter, dynamic paths, the fusion with an IMU offers clear advantages in accuracy and handling aggressive motions. This underscores the importance of context-specific system selection and the integration of corrective mechanisms for optimal odometry performance.
\section{Conclusion}\label{sec9}

This paper emphasizes the crucial role of LiDAR odometry in robotics, underlining its profound influence on perception and navigation. Our survey covers almost all recent LiDAR odometry advancements, delineating their strengths and weaknesses. The versatility of LiDAR odometry is evident, especially in environments with unreliable GPS, making it essential for robotic navigation and mapping. Furthermore, this paper addresses remaining challenges in LiDAR odometry discusses potential improvements and future directions in the field, and introduces a variety of datasets and evaluation metrics.

While a wealth of LiDAR odometry literature is available, unfortunately, there is no one-size-fits-all solution. LiDAR odometry involves a trade-off between resources and performance, requiring users to carefully consider these factors based on their specific application requirements and available resources. For low computational, especially with low-power single-board computers, a LiDAR-only approach may be optimal in well-defined environments. Integrating an IMU in a loosely-coupled fashion can enhance results without significantly increasing computational demands. A tightly-coupled multiple-sensor approach is advisable for applications demanding high accuracy across various environments. Combining LiDAR with an IMU is a balanced choice in general situations. Utilizing multiple LiDAR systems may be beneficial to address the narrow \ac{FOV} issue. Incorporating a camera can be advantageous in texture-limited scenarios. Those with greater computational resources can explore advanced capabilities offered by deep learning-based LiDAR odometry.

We anticipate the ongoing expansion of LiDAR odometry and believe that resolving the challenges through deep learning and multi-modal sensor fusion will pave the way for a general solution. Furthermore, we expect that the continuous development of both LiDAR sensors and odometry algorithms will lead to the emergence of even more accurate and robust odometry solutions in the future.

\section*{Acknowledgement}
This work was supported by the National Research Foundation of Korea(NRF) grant funded by the Korea government(MSIT) (No. RS-2023-00241758).

\bibliographystyle{sn-basic}
\bibliography{sn-bibliography, string-short}

\end{document}